%% file: paper-arXiv.tex
\documentclass{article}

\usepackage{hyperref}       % hyperlinks
\usepackage{url}            % simple URL typesetting
\usepackage{booktabs}       % professional-quality tables
\usepackage[margin=1in]{geometry}
\usepackage{fullpage}
\usepackage{graphicx}
\usepackage{amsmath}
\usepackage{amssymb}
\usepackage{amsthm}
\usepackage{float}
\usepackage{natbib}
\usepackage{xcolor}
\usepackage{enumitem}
\usepackage[mathscr]{eucal}
\usepackage{xspace}

\newcommand{\cX}{{\mathcal X}}
\newcounter{note}[section]

\def\ie{\mathrm{IE}}
\def\q{q}
\def\u{u}

\def\f{\mathsf{f}}
\def\t{\mathsf{t}}

\def\p{\mathsf{tpr}}
\def\n{\mathsf{fpr}}
\def\pipeline{\mathscr{P}}
\def\bypass{\mathsf{bypass}}
\newcommand{\argmax}{\mathrm{argmax}{}\xspace}
\newcommand{\argmin}{\mathrm{argmin}{}\xspace}

\newtheorem{thm}{Theorem}[section]
 \newtheorem{lem}[thm]{Lemma}
 \newtheorem{clm}[thm]{Claim}
 \newtheorem{obsr}[thm]{Observation}

 \newtheorem{defn}{Definition}[section]
 
 \newtheorem{rem}{Remark}

\title{Multi Stage Screening: Enforcing Fairness and Maximizing Efficiency in a Pre-Existing Pipeline}

\author{Avrim Blum\thanks{Toyota Technological Institute at Chicago (TTIC). Emails: \texttt{\{avrim, kevin, vakilian\}@ttic.edu}. This work was supported in part by the National Science Foundation under grants CCF-1815011 and CCF-1934843, and by the Simons Foundation under the Simons Collaboration on the Theory of Algorithmic Fairness.} \and Kevin Stangl\footnotemark[1] \and Ali Vakilian\footnotemark[1]}
\begin{document}
\date{}
\maketitle
\begin{abstract}
Consider an actor making selection decisions (e.g., hiring) using a series of classifiers, which we term a  {\em sequential screening process}.
The early stages (e.g. resume screen, coding screen, 
phone interview) filter out some of the applicants, and in the final stage an expensive but accurate test (e.g. a full interview) is applied to those individuals that make it to the final stage. 
Since the final stage is expensive, if there are multiple groups with different fractions of positives in them at the penultimate stage (even if a  slight gap), then the firm may naturally only choose to apply the final (interview) stage solely to the
highest precision group
%group of highest precision,  
which would be clearly unfair to the other groups.  
Even if the firm is required to interview all those who pass to the final round, the tests themselves could have the property that qualified individuals from some groups pass more easily than qualified individuals from others.
%Given these concerns, 

Accordingly, we consider requiring Equality of Opportunity (qualified members of each group have the same chance of reaching the final stage and being interviewed).
We then examine the goal of
maximizing quantities of interest to the decision maker subject to this constraint,
%The action space of our algorithm is modifying how we promote individuals through the process,  base based on performance at the previous stage.  d on performance at the previous stage.
via  modification of the probabilities of promotion through the screening process at each stage based on performance at the previous stage.

We exhibit algorithms for satisfying Equal Opportunity over the selection process and maximizing precision (the fraction of interviews that yield qualified candidates) as well as linear combinations of precision and recall (recall determines the number of applicants needed per hire) at the end of the final stage. 
We also present examples showing that the solution space is non-convex, which motivate our combinatorial exact and (FPTAS) approximation algorithms for maximizing the linear combination of precision and recall. 
Finally, we discuss the `price of' adding additional restrictions, such as not allowing the decision-maker to use group membership in its decision process. 
%\avnote{Do we discuss the price of fairness?}
\end{abstract}
\input{content}
\bibliography{mybib.bib}
\bibliographystyle{plainnat}
\appendix
\input{missing-proofs}

\input{appendix}

\end{document}

%% file: content.tex
\section{Introduction}
Consider what we will term \textit{sequential screening processes}. 
In this setting a decision maker (e.g. a company seeking to hire applicants) makes a decision, 
like hiring, by using a sequence of intermediate decision-making steps that each filter out some candidates, in order to ideally produce a pool of mostly qualified candidates at the final step. 

We assume some people are truly qualified for the position being filled, and we call them positive examples, and others are truly unqualified and we call them negative examples. And then the various intermediate steps have different probabilities of qualified/unqualified applicants passing each step, which could be different for different demographic groups.  We also assume that the final (interview) stage of the process is particularly expensive for the decision-maker, and reveals the true label of the applicant.

%%%%
%Imagine a two-step filtering and selection process to hire employees. In the first stage, the Learner has access only to resumes of individuals from group $A$ and group $B$.  Using the resumes to filter out un-promising applicants, the Learner then selects a subset to receive an interview, which is presumed to be perfectly accurate in selecting candidates to proffer offers.

%However, the stages differ in their cost structures for the firm. Scanning resumes via an automated system is cheap and its cost is negligible, but the interviewing stage is costly. In response, the firm will be very attentive to selecting only the most promising individuals to promote for interviewing, in order to minimize the expected number of interviews required to discover $k$ qualified candidates (verified by passing the final interview). Assume that the final interview can perfectly distinguish true positives from true negatives.

%In contrast to the standard PAC learning set-up,
%The goal of the learner is not to learn an accurate model for the sake 
%of learning an accurate model.
%An accurate model is only helpful if it can identify %enough
%qualified individuals. 

To illustrate a concern that could arise in this setting, suppose there are two demographic groups $A$ and $B$, and just one test $t$ in the screening process prior to the final stage.  Suppose that test $t$ and the underlying base rates of the two groups have the property that
%individuals are interviewed only if $t(x)=1$.
%For example, consider two groups $A$ and $B$ and a test $t$ such that
%with equal base rates and some $x$ and $x'$ identical other than group membership but such that 
$P(y=1|t(x) =1, x \in A)  \geq P(y=1 |t(x)=1, x \in B ) + \epsilon$ for some $\epsilon > 0$.  That is, the pool of group-$A$ applicants who pass the test has a higher fraction of positive examples than the pool of group-$B$ applicants who pass the test.
%; the probability that a randomly selected example from group $A$ has `positive' outcome (i.e., will pass the final interview stage') is slightly higher than the one for a randomly selected example from group $B$.
Since the cost of final interviews is assumed to be high, in this case a rational decision maker would be sensitive to even a small $\epsilon$ gap, in order to minimize the expected number of interviews made per hire. 
In particular, small gaps between these groups in the population would be amplified in that
the rational decision-maker 
would then choose not to promote \textit{any} individuals from group $B$ to the final interview round, which clearly violates common sense fairness norms. 
There is empirical evidence that similar phenomenon occurs in real world settings, when employers have limited information~\citep{bertrand2004emily}.

A second concern is that even if the decision-maker interviews all individuals who make it to the final round (and more generally, at each level promotes all individuals who pass the test to the next round), the tests themselves could have the property that qualified individuals from some groups pass them more easily than qualified individuals from others.  So, in the end, a qualified individual from one group might have a much lower chance of making it to the final interview round than a qualified individual from another.

%%%%
Because of fairness violations of this kind, we consider a regulator that requires the screening process to satisfy Equal Opportunity~\citep{hardt2016equality}, that is, qualified individuals of each group have the same chance of receiving an interview. 
This requirement motivates
%brings up
the 
%question
problem
of how to satisfy such a condition in the most efficient way, minimizing the number of interviews needed per successful hire as well as the number of overall applicants needed to enter the screening process per hire. 
This is the question we address in our paper.

We assume that the tests themselves and their order in the process are fixed beforehand and the action space of the firm (of our algorithm) is solely modifying how individuals move through the pipeline in response to their test outcomes (the promotion policy).
More specifically, for each test, we need to decide the probability that an individual from a given group who passes or fails the test should continue on to the next stage.
One can satisfy the fairness requirement with simple promotion policies (such as promoting all individuals regardless of whether they pass or fail each test), but the tension is how to do so in a way that results in a useful process.

This captures the scenario of performing  modifications to pre-existing screening systems (the test themselves are fixed) in order to respond to fairness issues. 
We assume we are given, for each test, its statistical properties for each group (the probability that a random qualified or unqualified individual will pass the test).\footnote{If we were to design a socio-technical system from first principles using the insights of machine learning research, we might seek to design tests that are ideally more robust to group difference and still predictive, however such a re-design process could be costly and slow. 
In a world of limited resources, re-purposing pre-existing tests to be more fairness aware in a timely manner and still maintaining effectiveness is necessary.} 

\subsection{Our Results}
We study how to implement the fairness requirement of Equal Opportunity in this sequential screening setting and what method of implementing it would achieve a high efficiency.  
One core result in our paper is that there is a solution that maximizes precision (minimizes the number of interviews needed per successful hire) subject to maintaining Equal Opportunity, that is given by promoting individuals from each group according to what we call the {\em opportunity ratio}. Moreover, it is possible to maximize overall precision subject to satisfying Equal Opportunity by a policy in which each level in the process satisfies Equal Opportunity individually (this property will not hold for the more general objective below). 
%is given by the Opportunity Ratio, in which we do not promote some fraction of group $A$ individuals who pass the first step of the pipeline in order to match the fraction of positives in group $B$ that pass the test.
%After the first step, we promote every sample who passes each test and promote no test-failers.

Then we consider the more general case of satisfying Equal Opportunity while maximizing a linear combination of precision and recall (1/precision is the expected number of interviews needed per successful hire, and 1/recall is proportional to the number of overall applicants needed to enter the screening process per hire). This problem is challenging because, as we show, the space of Equal-Opportunity solutions is non-convex. Moreover, the optimal way to use one test to optimize a linear combination of precision and recall may depend on %other tests in the system.
all other available tests.

Nonetheless, we are able to achieve an FPTAS for maximizing any linear combination of precision and recall, as well as an exact algorithm with running time that is `only' exponential in the number of levels $k$ and the number of the groups. This latter result relies on certain structural properties of optimal solutions that we develop in our analysis. 
Finally, we discuss extensions to our model such as requiring the screening process to be group-blind, and considering the requirement of satisfying Equalized Odds.  Unfortunately, the optimal fair group-blind policy may be much worse than the optimal fair group-aware policy. For example, in some cases it may require a policy that completely bypasses all the tests.
%In particular, in Section \ref{sec:single-policy} we show that when we are required to have a single promotion policy for all groups (group-blind), that requiring Equal Opportunity results in an a vacuous model in the worst case. 
 
%We additionally defer selected proofs to the appendix in order to enhance comprehension.
\subsection{Related Work}
Fairness in pipelines was initiated by~\citet{bower2017fair} and follow up work by~\citet{dwork2018fairness, dwork2020individual}. This paper
%model 
differs from~\citep{dwork2020individual} in several keys ways. 
We both use the word `pipelines' but our work is more focused on the specific case of hiring pipelines in which we are looking at the fairness of the final outcome for a given individual, drawn from the population, rather than considering the individual fairness~\citep{dwork2012fairness} of the cohort context to which one is assigned.
We do not consider cohort based scoring rules.

The structure of our model is very close to that of~\citet{downstream18}, but the objective in that work 
is jointly designing college admission and grading schemes that
satisfy Equal Opportunity over the admissions/college process \textit{and in particular} incentivize a rational employer to use a group blind hiring policy.
In contrast, our work considers maximizing precision or a linear combination of recall and precision while satisfying Equal Opportunity.

Another related work by~\citet{arunachaleswaran2020pipeline} is the idea of pipeline interventions. 
In that paper there is a wide pipeline with a finite number of states at time $t$ and the goal of the algorithm designer is to modify the transition probabilities from state to state in order to maximize a reward at the final step. 
This corresponds to efficiently allocating a government subsidy to aid dis-advantaged individuals, from the perspective of maximizing social welfare.

Intriguingly, the paper by~\citep{khalili2021fair} argues that Equal Opportunity is misaligned with fairness in screening allocation problems with a finite number of available items (think hiring a small number of engineers at a start-up vs accepting applicants for a credit card).
In our work, we do not focus on modeling a finite number of available positions (e.g., we are in the case with a larger number of available items).

Most closely related to our work is~\citet{mansourscreen}, in which there is noisy Bernoulli feedback in a hiring setting with sequential tests.
In contrast to our scenario, they assume both underlying candidate skill levels and test results are sampled independently from Bernoulli distributions. 
Furthermore, they allow hiring an applicant before the end of the pipeline (e.g., if you pass the first three of five tests and those tests have high signal, you may skip the next two tests).
In our model, we assume each stage of the process is memoryless (the probability of making it to stage 3 from stage 2 depends only on the result of the stage-2 test and group membership, and not the result of the stage-1 test) and we allow tests to be asymmetric (e.g., it could be that positive examples from a given group pass with probability 0.75 and negative examples pass with probability 0.5).
In our motivation, we model the initial tests as cheap while the ultimate interview is expensive and accurate, while in \citet{mansourscreen}, 
each test is equally accurate and costly and additionally they want to minimize the expected number of tests to hire a candidate. 
Consistent with our perspective, the authors exhibit an impossibility result arguing that satisfying Equal Opportunity requires group dependent thresholds if the tests have different noise rates.  
%In contrast to our work, they exhibit a greedy algorithm based on a random walk, but they are solving for a different objective (i.e. minimizing the expected number of tests) while we focus on the precision at the end of the pipeline and on the linear combination of precision and recall.

Additionally, there are connections between our work and classical economic discussions of statistical discrimination~\citep{arrow72, phelps72} in that both perspectives model disparities in outcomes that derive from strategic actors making decisions to allocate
goods differently based on perceived differences in predicted outcomes (termed statistical discrimination). 
Our models do not capture taste based discrimination.

\subsection{Roadmap} %SHRINK shorter
In Section~\ref{model} we formally describe our model and present some examples that show key phenomena. 
In Section~\ref{maxprecsection} we prove and discuss our first main theorem, about how to maximize precision (at the end of the screening process)
subject to Equal Opportunity.

Then we consider the more general case of satisfying Equal Opportunity while maximizing a linear combination of precision and recall. This problem is challenging because, as we show in Section~\ref{sec:problem-statement-and-example}, the space of Equal-Opportunity solutions is non-convex. Moreover,  
how to effectively utilize a test may depend on all other available tests (Section~\ref{sec-test-nonmono}).
On the other hand, as we show in Section~\ref{sec:exact-alg}, the solution space does satisfy certain useful structural properties.
We then use these structural results to to achieve an exact optimal algorithm, and in Section~\ref{sec:fptas} to achieve an FPTAS for maximizing linear combination of precision and recall, as well as other functions of precision and recall.

Finally, in Section~\ref{sec-altmodels} we discuss extensions to our model such as requiring the screening process to be group-blind, and considering the requirement of satisfying Equalized Odds. 
%This work provides a method to implement the Equal Opportunity constraints in this type of screening process, and explains trade-offs of fairness variations in this model. 
%We defer selected proofs to the appendix in order to enhance comprehension.

\section{Preliminaries\label{model}}
Now we formally define our model and introduce some informative examples.
As mentioned above, the scenario to keep in mind is a stylized hiring process, consisting of a sequence of tests or interviews. 
Each candidate takes a test, and depending on their outcome on that test at that stage, is possibly promoted to the next stage of the screening process. 
We focus on modifying this promotion policy in response to satisfying the fairness constraints and achieving a high objective value or a low cost value.
This is a constrained optimization problem, with structure. 

\subsection{Definitions}
We use ${\mathcal X}$ to denote the set of demographic groups, and $X\in {\mathcal X}$ to denote a specific group.
We assume group membership is known to the algorithm, groups are disjoint, and an individual from group $X$ is promoted based on both their test performance and a promotion policy (defined below) for that corresponding group.
We assume individuals are either truly qualified or truly unqualified, and use label $y=1$ to denote a truly-qualified individual and label $y=0$ to denote a truly-unqualified individual. For each group $X$, let $q_X$ denote the base rate for that group, namely $\Pr(y=1 | x \in X)$.

\begin{defn}[{\bf Test Statistics}]
For each test $t$ and each group $X\in {\mathcal X}$, we define $\tau_{X1} := \Pr[t(x, y)=1 | y=1, x\in X]$ to be the probability a qualified candidate from group $X$ passes the test, and $\tau_{X0} := \Pr[t(x, y)=1 | y=0, x\in X]$ to be the probability an unqualified candidate from group $X$ passes the test. 
We assume all tests are {\em minimally effective} for all groups in that positive examples are more likely to pass than negative examples. More precisely, \begin{align}\label{eq:minimally-effective}
    \tau_{X1} > \tau_{X0}\ge 0
     \quad \forall X\in \mathcal X &&\text{(Minimal Effectiveness Property)}
\end{align}
Note that we assume that the probability of an individual passing a given test depends only on their true qualification $y$ and their group membership $X$. We also assume test statistics are given and known to our algorithm.  
\end{defn}
We use 
$\tau_{X1}^{j}$, $\tau_{X0}^{j}$ to denote the test statistics at \emph{stage j} of the interview process.
For convenience, we define $T_{X}^j = (\tau^j_{X1}, \tau^j_{X0})$ as useful shorthand to capture the test statistics at stage $j$ for group $X$. 
Note that the same test may have different effectiveness per group.
%, and we use this effectiveness to decide how to promote or screen out individuals as they move through the screening process. 

\begin{defn} [{\bf Post-Processing Modification}]
We would like to modify the outcomes of the tests in the screening process so 
that some fairness goal (to be specified later) is achieved at the end of the screening (i.e., in the final interview stage).
Further, we assume as part of the problem setting that the only `allowed' correction is to modify how 
candidates are promoted to the next stage.
The promotion probability of each candidate only depends on their group membership and performance at the {\em current} test (whether they passed or failed the test). 
Formally, for each group $X\in {\mathcal X}$, let $\pi_{X1}^j$ denote the probability a candidate $x \in X$ who passes the test at stage $j$ is promoted to stage $j+1$, and $\pi_{X0}^j$ the probability that a candidate who fails the test at stage $j$ is promoted to stage $j+1$.\footnote{Note, in general randomized promotion policies will be necessary to satisfy the fairness criteria.}
We describe a policy for a given stage $j$ as $\{(\pi_{X1}^j, \pi_{X0}^j)\}_{X\in \mathcal X}$.
\end{defn}

For instance, a naive fairness respecting solution is to simply ignore the tests and promote all examples to the end of the pipeline, i.e., $\{(\pi_{X1}^j=1,\pi_{X0}^j=1)\}_{X \in \cX, j \in [k]}$  where $k$ is the number of
tests in this screening process.
However, this would result in a useless process from the perspective of the decision maker. 
The most straightforward use of tests is to promote all who pass and none who fail, i.e., $\{(\pi_{X1}^j=1,\pi_{X0}^j=0)\}_{X \in \cX, j \in [k]}$. However, this might not satisfy 
required fairness properties. 
%\AV{For consistency, better to use $k$ instead of $K$.} 
We now formally describe the fairness properties we consider. 

%If the decision maker only wants to maximize precision, this could be a useful solution, but depending on the $\tau_{Bi}$, could result in no positives from group $B$ reaching the end of the screening process and causing fairness concerns, or the positives in group $B$ have considerably smaller chance to reach the interview stage compared to the positives in group $A$. We emphasize that for each group $X\in \mathcal X$ and each stage $i\in [k]$, $T_{Xi}$ are characteristics of the test $t$ in level $i$ and are given to us and cannot be modified. 

\begin{defn}[\bf Equal Opportunity and Equalized Odds~\citep{hardt2016equality}] 
Our paper primarily discusses two fairness notions, specifically {\em Equal Opportunity} and {\em Equalized Odds}.
The first notion, Equal Opportunity requires that the classifier have equal {\em True Positive Rates} for each group in the population. Equivalently, 
for a classifier $h$ and true labels $y$, $P(h(x) =1 | y(x)=1, x \in A) = P(h(x) =1 | y(x)=1, x \in B)$.
Equalized Odds is similar but it also requires that the {\em False Positive Rates} are equal; formally, $P(h(x) =1 | y(x)=0, x \in A) = P(h(x) =1 | y(x)=0, x \in B)$.
%\footnote{While these notions are somewhat deprecated in the fairness in machine learning literature, their simplicity in auditing and deployment mandates that they continue to be studied from a theoretical perspective. \avnote{not sure we want to keep it.}}
\end{defn}
In our problem, Equal Opportunity is motivated by a desire that qualified individuals should have the same shot at an interview regardless of their group membership.
%, and a concern that without any constraints, a rational decision-maker might only select individuals to interview from whichever group has highest precision at the last level. 
%otherwise would in a sense, over-fit, to small gaps in precision at the final layer between the groups and totally discard the positive examples from a certain group even though there is a relatively small gap between the two groups. 
In our problem, there is additionally a critical distinction between %Equal Opportunity/Equalized Odds
the fairness criteria (e.g. Equal Opportunity or Equalized Odds) being satisfied at the end
%holding over the entire 
pipeline and alternatively that requiring these criteria hold for every transition between stages as individuals move through the pipeline, a stronger notion.

Now that we have described the terms that characterize a problem instance and the action space of the algorithm, we describe the objective value that captures the usefulness of a screening process.
We term these multiple different objective functions `pipeline efficiency'.
 \begin{defn}\label{def-pipeline-eff}{\bf Pipeline Efficiency}
% In contrast to the above fairness notions, we also want to capture the utility of the screening process from the perspective of the firm deploying the model. 
 In our work we focus on two core notions of efficacy from the perspective of the firm deploying the screening process.
 {\em Interview efficiency} (equivalently, {\em precision}) is the fraction of candidates in the last round who are qualified, i.e., the fraction of interviews that lead to hires (or at least to job offers). 
 {\em Throughput efficiency} (equivalently, {\em recall}) is fraction of qualified candidates who make it to the final round, and determines the expected number of applicants needed to enter the pipeline to hire one candidate.
% This quantity is proportional to the inverse of the fraction of true positives who make it to the end of the pipeline (recall).
 %
 In this paper, we study cost functions that are functions of these two quantities only.
 %; interview complexity ($\propto$ 1/precision) and advertisement complexity ($\propto$ 1/recall).
 \end{defn}
 We model the last available test as highly discriminative but extremely expensive per each test utilization  and this is what motivates the interview efficiency. 
 In particular, if we assume that the $k$ stages prior to the interview round have zero or negligible cost per test, and there are many available candidates,
 then we presume that the goal of the firm is to maximize the interview efficiency (precision, at the final round). 

\subsection{Formal Problem Statement and Illustrative Examples}\label{sec:problem-statement-and-example}
Now, we combine the above into a formal statement. 
Given a screening process/pipeline $\pipeline$ with $k$ stages, this pipeline consists of a collection of disjoint groups $\mathcal{X}$ and tests statistics $T_X = (T_{X}^{1}, T_{X}^{2}, \dots T_{X}^{k} )$ for every group $X\in \mathcal{X}$. 

%The goal of the algorithm designer is to exhibit a method to find promotions policies $\{(\pi_{X1}^{1}, \pi_{X0}^{1}), \dots, (\pi_{X1}^{k}, \pi_{X0}^{k})\}_{X\in \mathcal{X}}$ denoted as $\pi$ such that the overall policy satisfies the relevant fairness notion (either at the end of the screening process or at the end of each stage) and maximizes the given pipeline efficiency.
The goal of the algorithm designer is to exhibit a method to find promotion policies $\{(\pi_{X1}^{j}, \pi_{X0}^{j})\}_{X\in \mathcal{X}, j\in [k]}$ denoted as $\pi$ such that the overall policy satisfies the relevant fairness notion (either at the end of the screening process or at the end of each stage) and maximizes the given pipeline efficiency.
Now we move into illustrative examples.

\paragraph{An illustrative one-stage example:} Consider a one-stage pipeline with test parameters $$((\tau_{A1},\tau_{A0}), (\tau_{B1},\tau_{B0})) = ((1,0.5), (0.8,0.5)).$$
Observe that the policy of promoting individuals if and only if they pass the test does not satisfy Equal Opportunity. 
Instead, two policies that satisfy Equal Opportunity are
$P=((\pi_{A1}, \pi_{A0}), (\pi_{B1}, \pi_{B0})) = ((0.8,0), (1,0))$ and policy $Q=((1,0), (1,1))$.
In words, the policy $P$ would promote all individuals who passed the test from group $B$, but would only promote $80\%$ of those from group $A$. 
This down-weighting of group $A$ would suffice to satisfy Equal Opportunity. 
In contrast, policy $Q$ promotes all individuals from group $A$ who pass the test and promotes everyone from group $B$, regardless of their test score.
In this example, $P$ is the optimal Equal Opportunity policy with respect to precision.

\paragraph{The set of policies satisfying Equal Opportunity is not convex:}
Interestingly, for a two stage pipeline with two groups, the set of policies satisfying Equal Opportunity is not convex.
Consider a pipeline with first level $T_{A}^{1} = (3/4, 0)$ and $T_{B}^{1} = (1/2, 1/2)$
and with second level $T_{A}^{2} = (1/2, 1/2)$. and $T_{B}^2 = (3/4, 0)$. 
Consider policy $P$ with ($P^1_{A} = (1, 0)$, $P^1_{B} = (1, 1)$)  and ($P^2_{A} = (1,1)$, $P^2_{B} = (1,0)$).
This policy has recall $3/4$ for each group and therefore satisfies Equal Opportunity.
Consider policy $Q$ with parameters 
($Q^1_{A} = (1, 0)$, $Q^1_{B} = (1, 1/2)$) and
($Q^2_{A} = (1,1)$, $Q^2_{B} = (1,1)$).
This policy also has the recall of $3/4$ for each group and therefore also satisfies Equal Opportunity. 
However, the average of these two policies denoted as $\pi$ is ($\pi^1_{A} = (1,0)$,  $\pi^1_{B} = (1,3/4)$),
while ($\pi^2_{A} = (1,1)$ , $\pi^2_{B} = (1,1/2)$).
The recall for group $A$ is still $\frac{3}{4}$, while the recall for group $B$ is $(\frac{1}{2} + \frac{1}{2} \cdot \frac{3}{4})(\frac{3}{4} + \frac{1}{4} \cdot \frac{1}{2}) = \frac{49}{64} \neq \frac{3}{4}$. 

Thus this convex combination of policies does not satisfy Equal Opportunity and therefore the set of Equal Opportunity promotion policies is not convex.

\paragraph{Requiring Equalized Odds at each level can significantly harm performance:} The above example also shows that requiring Equalized Odds at each level can significantly harm performance.  Notice that policy $P$ above satisfies Equalized Odds overall and has perfect precision and fairly high recall.  However, the only way to satisfy Equalized Odds at each level is to completely bypass both tests, which would be much worse for precision.

Interestingly, as we show below, requiring {\em Equal Opportunity} at each level does {\em not} harm precision relative to requiring it for the pipeline as a whole (though it can hurt recall).

\section{Maximizing Precision Subject to Equal Opportunity \label{maxprecsection}}
% \section{Interview Efficiency: Maximizing Precision and Satisfying Equal Opportunity
In this section, we exhibit a policy $\pi$ that maximizes precision at the end of the screening  process while satisfying Equal Opportunity over the entire process.
To do this, we prove that the optimal method for this objective is given by promoting individuals from
each group according to the {\em Opportunity Ratio} (which we will define shortly).  
%Note, for clarity, we first define and consider these notions for a screening process with a single test before the final interview, and then extend to the general case with $k$ stages.
%\KS{First, consider the notion of True Positive  and False Positive Rates, which we can write out explicitly as a function of the test parameters and our promotion policy.}

\begin{defn}
For a test $\tau$ and associated promotion policy $\{(\pi_{X1}, \pi_{X0})\}_{X\in \mathcal X}$, define $M_{X,\tau, \pi} := (\tau_{X1} \pi_{X1} + (1 - \tau_{X1}) \pi_{X0})$ and $N_{X,\tau, \pi} := (\tau_{X0} \pi_{X1} + (1 - \tau_{X0}) \pi_{X0})$. Note that $M_{X,\tau, \pi}$ and $N_{X,\tau, \pi}$ are the probabilities that a positive and respectively a negative example from group $X$ is promoted to the next level,
%{\em True Positive Rate} and the {\em False Positive Rate} of $\pi$ the probability 
and so will be important quantities for our analysis.

\end{defn}

%\KS{should we say something about k stages here?}
%Now we consider the interview efficiency for a screening process with one single test.
\begin{obsr}\label{obs:EOpp-cond}
For any single-stage policy $\{(\pi_{X1}, \pi_{X0})\}_{X\in \mathcal X}$ that satisfies Equal Opportunity for a test with parameters $\{(\tau_{X1}, \tau_{X0})\}_{X\in \mathcal X}$, there exists $M$ such that $M_{X,\tau, \pi} = M$ for every $X\in \mathcal X$.

Furthermore, for a $k$-stage screening process $\{\tau^i\}_{i\in [k]}$, a policy $\{(\pi_{X0}, \pi_{X1})\}_{X\in \mathcal X}$ is Equal Opportunity if there exists $M$ such that $\Pi_{i=1}^k M_{X,\tau^i, \pi^i} = M$ for every group $X \in {\mathcal X}$.
\end{obsr}
%These observations are identical to the definition of Equal Opportunity. 

% insert opportunity ratio and an example

\begin{obsr}\label{obsr:interview-eff-formula}
Recall that $q_X$ denotes the base rate for group $X$, and let $u_X = 1-q_X$. For a single-stage pipeline with test $\tau$ and promotion policy $\pi$,
%$\{(\pi_{X1}, \pi_{X0})\}_{X\in \mathcal X}$ and the pre-interview test $t$ with qualified and unqualified fractions $\{(\q_X, \u_X)\}_{X\in \mathcal X}$, 
the interview efficiency (i.e., precision) is equal to
\begin{align}\label{eq:interview-eff}
    \ie(q,u,\tau,\pi):= \frac{\sum_{X\in \mathcal X}\q_X M_{X,\tau, \pi}}{\sum_{X\in \mathcal X}\q_X M_{X,\tau, \pi} + \u_X N_{X,\tau, \pi}}.
\end{align}
Similarly, when we consider the extension to a $k$-stage pipeline, the interview efficiency is equal to
\begin{align}\label{eq:interview-eff-k}
    \ie(q,u,\tau,\pi):= \frac{\sum_{X\in \mathcal X}\q_X \prod_{i=1}^{k} M_{X,\tau^{i}, \pi^{i}}}{\sum_{X\in \mathcal X}\q_X \prod_{i=1}^{k} M_{X,\tau^{i}, \pi^{i}} + \u_X \prod_{i=1}^{k} N_{X,\tau^{i}, \pi^{i}}}.
\end{align}
\end{obsr}

Now, we formally define the policy given by the opportunity ratio as follows.
\begin{defn}[\textbf{Opportunity Ratio Policy}] Consider a screening process with $k$ stages. For each $X\in \mathcal{X}$, let $\rho_X := \Pi_{j\in [k]}(\tau^j_{X^*1}/\tau^j_{X1})$, where $X^* = \argmin_{X\in \mathcal X} \Pi_{j\in [k]}\tau^j_{X1}$.
The {\em Opportunity Ratio} policy, at the first stage for each $X\in \mathcal{X}$, promotes $\rho_X$ fraction of those who pass the test and none of those who fail the test. For the remaining stages $(i=2,3,...,k)$, the Opportunity Ratio policy fully trusts the result of the tests; a candidate is promoted to the next stage iff they pass the test at the current stage. Formally, for every $X\in \mathcal{X}, \pi^1_{X1} = \rho_X, \pi^1_{X0}=0$ and $\pi^i_{X1}=1, \pi^i_{X0} =0, \forall i\ge 2$. 
\label{def:OR}
\end{defn}

In the rest of this section, 
%we show that promoting according to this policy is the unique
we study the task of maximizing interview efficiency under different settings and fairness requirements.
\subsection{Maximizing Interview Efficiency subject to Equal Opportunity at the Final Stage} 
As a warm-up, we start with the simplest setting where the screening process has only one test before the interview stage.
%(final stage).
\begin{thm}[{\bf Opportunity Ratio Policy Maximizes Precision for Single-Stage %Screening
Process}]\label{thm:single-pipeline}
Let $t=((\tau_{A1},\tau_{A0}), (\tau_{B1},\tau_{B0}))$ be a test satisfying the minimally effectiveness property.  
The maximum precision policy satisfying Equal Opportunity is the opportunity ratio policy.
Moreover, for any group $X\in \mathcal X$, it is always sub-optimal to promote any candidates who failed the test (i.e., in any optimal policy, $\pi_{X0} = 0, \forall X\in \mathcal{X}$).
\end{thm}
\begin{proof}
First, for any policy $\pi$, we upper-bound the interview efficiency (i.e.,~precision) for a screening process with parameters $\q, \u, \tau$. To bound the interview efficiency, for each $X\in \mathcal X$, we lower-bound the False Positive Rate $N_{X,\tau,\pi}$ in terms of the True Positive Rate $M_{X,\tau, \pi}$. 
\begin{align}
    N_{X,\tau, \pi} = \tau_{X0} \pi_{X1} + (1-\tau_{X0}) \pi_{X0}
    &= \tau_{X0}(\pi_{X1}-\pi_{X0}) + \pi_{X0} \nonumber\\
    % &\ge \tau_{X0} (\pi_{X1}-\pi_{X0}) + \frac{\tau_{X0}}{\tau_{X1}} \pi_{X0} &&\rhd\text{by Eq.~\eqref{eq:minimally-effective}, $\forall X\in \mathcal X$, $\tau_{X1} > \tau_{X0} \ge 0$} \nonumber\\
    &\ge \frac{\tau_{X0}}{\tau_{X1}} \big(\tau_{X1}(\pi_{X1}-\pi_{X0}) + \pi_{X0}\big) &&\rhd\text{by Eq.~\eqref{eq:minimally-effective}, $\forall X\in \mathcal X$, $\tau_{X1} > \tau_{X0} \ge 0$} \nonumber\\
    &= \frac{\tau_{X0}}{\tau_{X1}}\cdot M_{X,\tau, \pi} \label{eq:X-bound}
\end{align}
By Equal Opportunity of $\pi$ and employing Eq.~\eqref{eq:X-bound} in the formula for the interview efficiency, Eq.~\eqref{eq:interview-eff},
\begin{align}
    \ie(\q, \u, \tau, \pi) 
    = \frac{\sum_{X\in \mathcal X}\q_X M_{X,\tau, \pi}}{\sum_{X\in \mathcal X}\q_X M_{X,\tau, \pi} + \u_X N_{X,\tau, \pi}} &\le \frac{\sum_{X\in \mathcal X}\q_X M_{X,\tau, \pi}}{\sum_{X\in \mathcal X}(\q_X + \u_X\cdot \frac{\tau_{X0}}{\tau_{X1}}) M_{X,\tau, \pi}} &&\rhd\text{by Eq.~\eqref{eq:X-bound}} \nonumber\\
    &= \frac{\sum_{X\in \mathcal X}\q_X}{\sum_{X\in \mathcal X}(\q_X + \u_X\cdot \frac{\tau_{X0}}{\tau_{X1}})} &&\rhd\forall X\in \mathcal{X}, M_{X,\tau, \pi} = M \label{eq:upper-bound-interview-eff}
\end{align}
Note that the inequalities are tight when $\pi_{X0} = 0$ for all $X\in \mathcal X$. 

Next, we show that the {\em opportunity ratio} policy satisfies Equal Opportunity and achieves the bound in Eq.~\eqref{eq:upper-bound-interview-eff}. 
In the opportunity ratio policy $\pi^*$, only a $(\frac{\tau_{X^*1}}{\tau_{X1}})$-fraction of candidates in group $X$ who pass the test $t$ (picked uniformly at random) are promoted to the next stage. In other words, for any group $X\in \mathcal X$, we set $\pi^*_{X1} = \frac{\tau_{X^*1}}{\tau_{X1}}, \pi^*_{X0}=0$. Then,
\begin{align*}
    \ie(\q, \u, \tau, \pi^*) 
    = \frac{\sum_{X\in \mathcal X}\q_X M_{X,\tau, \pi^*}}{\sum_{X\in \mathcal X}\q_X M_{X,\tau, \pi^*} + \u_X N_{X,\tau, \pi^*}} \nonumber 
    &= \frac{\sum_{X\in \mathcal X}\q_X \tau_{X1} (\frac{\tau_{X^* 1}}{\tau_{X 1}})}{\sum_{X\in \mathcal X} \q_X \tau_{X1} (\frac{\tau_{X^* 1}}{\tau_{X 1}}) + \u_X \tau_{X0}(\frac{\tau_{X^* 1}}{\tau_{X 1}})} \\
    &= \frac{\sum_{X\in \mathcal X}\q_X}{\sum_{X\in \mathcal X}(\q_X + \u_X\cdot \frac{\tau_{X0}}{\tau_{X1}})} %&&\rhd \text{by~\eqref{eq:minimally-effective}, $\tau_{X^*1}>0$}
\end{align*}
Hence, $\pi^*$ is an equal opportunity policy with the maximum interview efficiency for any screening process with parameters $\q, \u, \tau, \pi$. 
%\AV{Note that our analysis does not use the fact that $q_A + u_A + q_B + u_B =1$. We may need to exploit this property further in the proof for multi-stage pipelines.}
\end{proof}
\begin{rem}
Note that any policy $\pi$ where for each $X\in {\mathcal X}$, $\pi_{X1} = \eta \cdot \pi^*_{X1}, \pi_{X0} = 0$ for a constant $\eta<1$ also satisfies the Equal Opportunity and maximizes the interview efficiency objective (i.e., precision). However, $\pi^*$ has a strictly higher {\em recall}. 
\end{rem}
%\subsection{Maximizing Interview Efficiency in Multi-Stage Processes}
Next, we state our result for the general setting in which there are multiple stages and multiple groups in the screening process. The proof of the theorem is similar to the single test version and is deferred to Appendix~\ref{sec:precision-max-proof}. %of the Appendix. 
\begin{thm}[{\bf Multi-Stage Screening Process}]\label{thm:multi-eq-opp-prec}
Consider a $k$-stage screening process whose all tests are minimally effective. The maximum interview efficiency policy satisfying Equal Opportunity is the Opportunity Ratio policy and has interview efficiency equal to $\frac{\|q\|_1}{\|q\|_1 + \sum_{X\in \mathcal X}\u_X \Pi_{i=1}^k(\tau^i_{X0}/\tau^i_{X1})}$.
\end{thm}

\subsection{Maximizing Interview Efficiency Subject to Equal Opportunity at the End of Each Stage} 
 Here, we consider the setting in which the goal is find a policy that maximizes interview efficiency and satisfy Equal Opportunity at the end of each stage---{\em not only at the interview stage}. 
Following Theorem~\ref{thm:multi-eq-opp-prec}, the maximum interview efficiency in this setting is at most $\|q\|_1/(\|q\|_1 + \sum_{X\in \mathcal X}\u_X \Pi_{i=1}^k\frac{\tau^i_{X1}}{\tau^i_{X0}})$. Next, we show that the following slightly modified opportunity ratio policy $\pi$ that satisfies Equal Opportunity at the end of each stage maximizes the interview efficiency. The policy $\pi$ applies the opportunity ratio at each stage of the pipeline.
\begin{align*}
    \pi^i_{X0} = 0, \pi^i_{X1} = \frac{\tau^i_{X^*_i 1}}{\tau^i_{X 1}} &&\forall i\in [k], X\in {\mathcal X}, \text{ where $X^*_i := \argmin_{X\in \mathcal X} \tau^i_{X1}$}
\end{align*}
Again, it is straightforward to verify that $\pi$ satisfies the Equality of Opportunity. Moreover, %the interview efficiency of $\pi$ is
\begin{align*}
    \ie(q, u, \tau, \pi)     
    = \frac{\sum_{X\in \mathcal X}\q_X M_{X,\tau, \pi}}{\sum_{X\in \mathcal X}\q_X M_{X,\tau, \pi} + \u_X N_{X,\tau, \pi}} %\\
    &= \frac{\sum_{X\in \mathcal{X}}q_X \Pi_{i\in [k]} \tau^i_{X^*_i1}}{\sum_{X\in \mathcal{X}}q_X \Pi_{i\in [k]} \tau^i_{X^*_i1} + \sum_{X\in \mathcal X}\u_X \frac{\tau^i_{X^*_i 1} \tau^i_{X 0}}{\tau^i_{X 1}}} \\
    &= \frac{\|q\|_1}{\|q\|_1 + \sum_{X\in \mathcal X}\u_X \Pi_{i=1}^k\frac{\tau^i_{X0}}{\tau^i_{X1}}}
\end{align*}
The only difference compared to the policy of Theorem~\ref{thm:multi-eq-opp-prec} is that in the former policy the recall can be higher.
\begin{rem}
Adding the condition to satisfy the Equality of Opportunity at the end of each stage does not harm interview efficiency. However, this condition may decrease the recall of the optimal policy.  
\end{rem}

\section{Pipeline Efficiency: Maximizing Linear Combinations of Precision and Recall}\label{sec:linear-objective}
%Preliminaries for Linear Combination of Recall/Precision} 
Now we shift our focus to exhibiting a promotion policy that satisfies Equal Opportunity and
maximizes a linear combination of precision and recall given by the positive weight $\alpha \in \mathbb{R}_{\ge 0}$; $f_{\alpha}(\pi):=(1-\alpha)\cdot \mathrm{recall}(\pi) + \alpha\cdot\mathrm{precision}(\pi)$.
As in Definition \ref{def-pipeline-eff}, higher precision corresponds to higher interview efficiency, and higher recall corresponds to higher throughput efficiency. 

We start with a simple $2$-approximation algorithm for maximizing any given linear  of precision and recall.
\begin{thm}[\bf Approximation Algorithm for Linear Combination of Precision and Recall]
There exists a  polynomial time 2-approximation algorithm for maximizing any linear combination of precision and recall. 
\end{thm}
\begin{proof}
Note that the policy that bypasses all tests is an Equal Opportunity policy and maximizes recall---it achieves recall equal to one.
Moreover, by Theorem~\ref{thm:multi-eq-opp-prec}, the Opportunity Ratio is an Equal Opportunity policy maximizing precision. Hence, the better of the ``bypassing all tests'' policy and the Opportunity Ratio policy is a $2$-approximation of any given linear combination of precision and recall. 
\end{proof}

In order to obtain better performance for maximizing linear combinations of precision and recall, we develop structural properties of optimal solutions, and then use them to get an exact algorithm with running time that is exponential only in $k$ and the number of groups.
Additionally, by a dynamic programming approach we exhibit a {\em fully polynomial time approximation scheme (FPTAS)}.
%as well as an FPTAS.

One challenge is that as shown in Section~\ref{sec:problem-statement-and-example}, the space of Equal Opportunity solutions is non-convex. Another is that as shown in Section \ref{sec:optnotopt} below, Opportunity Ratio is no longer optimal, and as shown in Section \ref{sec-test-nonmono} below, there exists no function ranking the efficacy of tests solely based on their statistics.  
%explain the challenges non-convexity/example 5

We begin by presenting the examples mentioned above, and then developing the structural properties we will use.

\subsection{Illustrative Examples}
\subsubsection{Opportunity Ratio not Optimal for Linear Combination of Precision and Recall}\label{sec:optnotopt}
In the previous sections, our 
key algorithmic strategy is to use the Opportunity Ratio to re-weight the promotion  policy. 
Since this policy satisfied Equal Opportunity and maximized precision (among Equal Opportunity policies), if our objective is to only maximize precision, then the Opportunity Ratio is sufficient.
Now we exhibit an example where the Opportunity Ratio solution is not optimal when maximizing any linear combination of precision and recall when there is any nonzero weight on recall.  Specifically, in this example there is an alternative policy with the same precision as the Opportunity Ratio solution but strictly higher recall.
%In a sense, this is obvious, since the opportunity ratio is focused on maximizing precision (recall for the opportunity ratio policy we do not promote any individuals who fail a test).

 Consider a pipeline with $T_{A}^{1} = (3/4, 0)$ and $T_{B}^{1} = (1/2, 1/4)$.
In the second stage, $T_{A}^{2} = (1/2, 1/4)$  and $T_{B}^{2} = (3/4, 0)$.
Consider policy $P$:  ($P_{A}^{1} = (1, 0)$ and $P_{B}^{1} = (1, 1)$, while $P_{A}^{2} = (1,1)$ and  $P_{B}^{2} = (1,0)$.

This policy has recall $3/4$ and precision $1$ for each group and therefore satisfies Equal Opportunity. 
Thus if our objective here is maximize the average of precision and recall, this policy has objective function value $7/8$.
In  contrast, the Opportunity Ratio policy as given in Definition \ref{def:OR} is $P_{A}^{1} = (1, 0)$,$P_{B}^{1} = (1, 0)$ and $P_{A}^{2} = (1,0)$, $P_{B}^{2} = (1,0)$ which reduces our recall to $\frac{3}{4} \cdot \frac{1}{2} = \frac{3}{8}$ while to precision is still $1$, for score of $\frac{11}{16}$.
Clearly this is a lower objective function score than the first policy.
%[ 3/4 0], [1/2, 1/4]

\subsubsection{Optimal Policy Non-Locality for Linear Combination of Precision and Recall \label{sec-test-nonmono}}
% Suppose we have one group in the population and want to optimize a linear combination of recall and precision---this is indeed a much simpler setting compared to the multi-group inputs with Equal Opportunity requirement that we eventually aim to solve). 
% A natural question is whether we can solve this problem with a simple greedy algorithm that makes local decisions in a single pass of the test statistics.
% We answer this question in the negative by exhibiting an example pipeline with test statistics such that when two of three tests are available, using only the first test is strictly optimal, while when considering all three tests the strict optimum is the opposite of the previous case.
% This shows that to find an optimal policy that maximizes a given linear combination of precision and recall, it is not sufficient to follow a ranking function $r : \mathcal{T} \rightarrow \mathbb{R}$ of the tests in the pipeline that determine the ``efficacy'' of each test solely based on its statistics (i.e., for each level $i$, $r$ is a function of $\{(\tau_{X1}, \tau_{X0})\}_{X\in \mathcal{X}}$, where $\mathcal{T}$ is the collection of tests used in various stages of the pipeline. 

%\avnote{Proofread this section.}
Suppose we have one group in the population and want to optimize a linear combination of recall and precision. 
A baseline idea is whether we can solve this problem with a 
%simple
natural
greedy algorithm that makes local decisions in a single pass of the test statistics \footnote{In the related work by \cite{mansourscreen} the answer is in the affirmative, but their model is different and has uniform noise across true positives and true negatives.}.

We answer this question in the negative in
by exhibiting an example pipeline with test statistics such that when two of three tests are available, 
using only the first test is strictly optimal, while when all three tests are available, the optimum is instead to use the other two tests and not the first test.
This shows that an algorithm that maximizes a linear combination of precision and recall cannot simply assign separate scores to each test and then use only the highest-scoring tests.  
Our example is only for one group. 

The counterexample is as follows. 
The base-rate in the population is  $P(y=1) = 1/2$.
Consider test $t_1 = (1/2,0)$ and tests $t_2 =t_3= (1-\delta, 1/2)$ where $\delta=\frac{1}{100}$. The objective function is $f(\pi) = \frac{1}{3} \cdot  \mathrm{recall}(\pi) + \frac{2}{3} \cdot \mathrm{precision}(\pi)$. 
%Further, $\pi^{i}=[ \pi^i_{1}, \pi^i_{0}]$ is the promotion probabilities for those who pass/fail test $t_i$ (there is only one group in this example). 
In the following, 
%as we work with policies that either fully use each test (i.e., $\pi_1 =1, \pi_0 = 0$) or bypass it (i.e., $\pi_1 = \pi_0 = 1$), we 
let $f(t_{1})$ to denote the score of the policy that only promotes those who pass $t_1$ and bypasses all other tests
while $f(t_{2} t_{3})$ denotes bypassing $t_1$ and promoting individuals if and only if they pass tests $t_2$ and $t_3$.
In the Appendix~\ref{appendix-linear-combination-counter} we show while $f(t_1)$ is larger than any policy using $t_1$ and $t_2$ (possibly in fractions), $f(t_2 t_3)$ is strictly larger than any policy using $t_1, t_2$ and $t_3$ (again, possibly in fractions).

% Using test $t_1$ ($\pi^{1}=[1,0]$) and bypassing test $t_2$ has recall $=1/2$ and precision $=1$, so $f(t_1)=5/2$.
% If test $t_2$ is available besides test $t_1$ then using test $t_2$ while keeping $\pi^{1}= [1,0]$ can only harm the recall, so once we are committed to setting $\pi^{1} = [1,0]$ the optimal choice is to bypass $t_2$; set $\pi^{2}=[1,1]$.

% Now, consider if a third test $t_3=t_2$ (e.g. an identical copy of $t_2$) is also available. 
% Then bypassing test $t_1$ and using test $t_2$ followed by test $t_3$ ($\pi^{2}=\pi^{3}=[1,0])$ has recall $(1-\delta)^2$.
% Similarly, this has precision $\frac{(1-\delta)^2}{(1-\delta)^2 + 1/4}$.
% So if $\delta = 1/100$, then 
% \[f(t_2 t_3)= (\frac{99}{100})^2 + 2\frac{(99/100)^2}{(99/100)^2 + 1/4} > 2.57 > \frac{5}{2} = f(t_1) \]
% Observe that $f(t_1 t_2 t_3) < f(t_1)$.
% In the Appendix~\ref{appendix-linear-combination-counter} we will further elaborate why $f(t_2 t_3)$ is the optimal policy for these problem parameters.

\subsection{An Exact Algorithm}\label{sec:exact-alg}
%Maximizing Linear Combination of Precision and Recall}
%\subsubsection{Properties of Pareto Optimal Policies}
%\avnote{starting this section, we switch to the case in which the equal opportunity is required only at the interview stage.}
In this section, we give an exact algorithm for maximizing any given linear combination of precision and recall subject to satisfying Equal Opportunity by the end of the screening process. 
%To recall, Equality of Opportunity in a $k$-stage screening process requires that for every pair of groups $X, Y \in \mathcal{X}$, $\Pi_{i\in [k]} M_{X,\tau^i, \pi^i} = \Pi_{i\in [k]} M_{Y,\tau^i, \pi^i}$.

First we show that for any $k$-stage screening process over a population specified by a collection of groups $\mathcal{X}$, there exists a set of Equal Opportunity policies $\mathcal{P}_{k, \mathcal{X}}$ that {\em weakly Pareto dominate} (w.r.t.~precision and recall) any policy satisfying Equal Opportunity. In particular, we show that each policy $\pi :=(\pi^1, \cdots, \pi^k)\in {\mathcal P}_{k, {\mathcal X}}$ has the following structure, $(1-\pi^i_{X1})\pi^i_{X0} = 0, \forall i\in [k], X\in \mathcal X$.
% \begin{align*}
%     (1-\pi^i_{X1})\pi^i_{X0} = 0, &&\forall i\in [k], X\in \mathcal X
% \end{align*}

\begin{defn}[{\bf Pareto Dominant Policy}]\label{def:pareto-dominant}
For a given screening process, a policy $\pi$ {\em weakly} Pareto dominates a policy $\tilde\pi$ w.r.t.~precision and recall iff, $\mathrm{recall}(\pi) \ge \mathrm{recall}(\tilde{\pi})$ and $\mathrm{precision}(\pi) \ge \mathrm{precision}(\tilde{\pi})$.
% \begin{align*}
%     &\mathrm{recall}(\pi) \ge \mathrm{recall}(\bar{\pi}),  &\mathrm{precision}(\pi) \ge \mathrm{precision}(\bar{\pi})
% \end{align*}
Moreover, $\pi$ {\em strictly} Pareto dominates $\tilde \pi$ if at least one of the above inequalities holds strictly. 

Furthermore, a set of policies $\mathcal P$ weakly Pareto dominates a policy $\tilde\pi$ w.r.t. precision and recall iff there exists a policy $\pi\in \mathcal P$ that $\pi$ weakly Pareto dominates $\tilde \pi$.
\end{defn}
% First, we set up some notations. For any group $X\in \{A, B\}$,
% \begin{align*}
%     &M^X_{\tau, \pi} := \tau_{X1}\pi_{X1} + (1-\tau_{X1})\pi_{X0}
%     &N^X_{\tau, \pi} := \tau_{X0}\pi_{X1} + (1-\tau_{X0})\pi_{X0} 
% \end{align*}
% Then, the Equality of Opportunity requires that 
% \begin{align}\label{eq:eqopp}
% \Pi_{i\in [k]} M^A_{\tau^i, \pi^i} = \Pi_{i\in [k]} M^B_{\tau^i, \pi^i}.
% \end{align}

% Recall that the precision of any policy $\pi$ is defined as follows. 
% \begin{align}\label{eq:precision}
%     \frac{\sum_{X\in \mathcal X} q_X \Pi_{i\in [k]} M^X_{\tau^i, \pi^i}}{\sum_{X\in \mathcal X} q_X \Pi_{i\in [k]} M^X_{\tau^i, \pi^i} + u_X \Pi_{i\in [k]} N^X_{\tau^i, \pi^i}}
% \end{align}
\begin{lem}\label{lem:pi-cond}
For any $k$-stage screening policy that satisfies the ``minimally effectiveness'' property, the set of Equal Opportunity policies in $\mathcal{P}:=\{\pi \in [0,1]^{2|\mathcal{X}|k} : (1 - \pi^i_{X1}) \pi^i_{X0} = 0, \forall X\in \mathcal{X}, i\in [k]\}$ weakly Pareto dominates all equal opportunity policies w.r.t.~precision and recall.
% \begin{align}\label{eq:pareto-front}
%     (1 - \pi^i_{X1}) \pi^i_{X0} = 0, \quad \forall i\in [k].
% \end{align}

In other words, any equal opportunity policy violating $(1 - \pi^i_{X1}) \pi^i_{X0} = 0$ for a group $X\in \mathcal X$ and a stage $i\in [k]$ is weakly Pareto dominated by ${\mathcal P}$. %(i.e., a policy that satisfies Eq.~\eqref{eq:pareto-front}). 
\end{lem}
\begin{proof}
First, we show that in any policy $\pi$ which is not strictly Pareto dominated (w.r.t.~precision and recall), $\pi^i_{X1}>0$ for every $X\in \mathcal X, i\in [k]$. Hence, we can only consider policies $\pi$ where $\pi_{X1}>0$ for all $X\in \mathcal{X}$. The proof of the following claim is deferred to Appendix~\ref{sec:linear-objective-proofs}. 
\begin{clm}\label{clm:non-zero-pi-one}
Consider a $k$-stage screening process whose tests satisfy the ``minimal effectiveness'' property.
In any optimal policy of this screening process that satisfies Equal Opportunity, for all $X\in \mathcal X$ and $i\in [k]$, $\pi^i_{X1} >0$. 
% In other words, any Equal Opportunity policy $\pi$ with $\pi^i_{X1} =0$ for a group $X\in \mathcal X$ and a stage $i\in [k]$ is strictly Pareto dominated by an Equal Opportunity policy $\tilde{\pi}$ such that $\tilde{\pi}^i_{X1}>0$ for all $X\in \mathcal X$ and $i\in [k]$. 
\end{clm}
Now, for the sake of contradiction, suppose that there exist a level $i\in [k]$ and a group $X\in \mathcal X$ such that $\pi^i_{X0}> 0$ and $\pi^i_{X1} < 1$. 
Note that w.l.o.g., we can assume that $\tau^i_{X1} <1$; otherwise, by setting $\pi_{X0} =0$, the recall of the policy does not decrease and the precision strictly increases. Hence, there exist $\epsilon_1, \epsilon_0 >0$ such that $\tau^i_{X1} \epsilon_1 - (1-\tau^i_{X1}) \epsilon_0 = 0$ where either $(\epsilon_1 = 1-\pi_{X1}, \epsilon_0 \le \pi_{X0})$ or $(\epsilon_1 \le 1- \pi_{X1}, \epsilon_0 = \pi_{X0})$. 

We define a new policy $\tilde{\pi}$, which differs from $\pi$ only in level $i$ of group $X$, as follows: $\tilde{\pi}^i_{X1} = \pi^i_{X1} +\epsilon_1$ and $\tilde{\pi}^i_{X0} = \pi^i_{X0} - \epsilon_0$.
% \begin{align*}
%     \tilde{\pi}^i_{X1} &= \pi^i_{X1} +\epsilon_1, &&\tilde{\pi}^i_{X0} = \pi^i_{X0} - \epsilon_0 \\
%     \tilde{\pi}^j_{Y1} &= \pi^j_{Y1}, 
%     &&\tilde{\pi}^j_{Y0} = \pi^j_{Y0} \quad \text{if $(Y\neq X) \vee (j\neq i)$}
% \end{align*}
%since for every $i\in [k]$, $N^X_{\tau^i, \pi^i} >0$, the new policy which replaces $\pi^i_{X1}$ with $\pi^i_{X1} +\epsilon_1$ and replaces $\pi^i_{X0}$ with $\pi^i_{X0} - \epsilon_0$, has higher precision which contradicts the optimally of policy $\pi$.
Next, we show that $N_{X,\tau^i, \tilde{\pi}^i} < N_{X,\tau^i, \pi^i}$.
\begin{align*}
    N_{X,\tau^i, \tilde{\pi}^i}
    &= \tau^i_{X0} \tilde{\pi}^i_{X1} +(1-\tau^i_{X0}) \tilde{\pi}^i_{X0} \\
    &= \tau^i_{X0} (\pi^i_{X1} + \epsilon_1) +(1-\tau^i_{X0}) (\pi^i_{X0} - \epsilon_0) \\
    &= \tau^i_{X0}\pi^i_{X1} +(1-\tau^i_{X0})\pi^i_{X0} + (\tau^i_{X0}\epsilon_1 + \tau^i_{X0}\epsilon_0 - \epsilon_0) \\
    &= \tau^i_{X0}\pi^i_{X1} +(1-\tau^i_{X0})\pi^i_{X0} + (\tau^i_{X0}\epsilon_1 + \tau^i_{X0}\epsilon_0 - \tau^i_{X1}\epsilon_1 - \tau^i_{X1}\epsilon_0) &&\rhd\text{since $\epsilon_0 = \tau^i_{X1} (\epsilon_0 + \epsilon_1)$} \\
    &< \tau^i_{X0}\pi^i_{X1} +(1-\tau^i_{X0})\pi^i_{X0} &&\rhd\text{since $\tau^i_{X0} < \tau^i_{X1}$} \\
    &= N_{X,\tau^i, \pi^i}
\end{align*}
%This implies that $\Pi_{j=1}^k N^X_{\tau^j, \pi^j} < \Pi_{j=1}^k N^X_{\tau^j, \tilde{\pi}^j}$. 
Further, since $\tau^i_{X1} \epsilon_1 - (1-\tau^i_{X1}) \epsilon_0 = 0$, $\tilde{\pi}$ satisfies Equal Opportunity and has the same recall as $\pi$. Moreover, since $N_{X, \tau^i, \tilde{\pi}^i} < N_{X, \tau^i, \pi^i}$ and for all $j\in [k]\setminus \{i\}$, $N_{X, \tau^j, \pi^j} \ge0$, $\Pi_{j=1}^k N_{X, \tau^j, \pi^j} \le \Pi_{j=1}^k N_{X, \tau^j, \tilde{\pi}^j}$. Hence the precision of $\tilde{\pi}$ is not less than the one of $\pi$. This contradicts the strict Pareto optimally of policy $\pi$. Thus the statement holds and for any level $i\in [k]$ and any group $X\in \mathcal X$, $(1-\pi^i_{X1})\pi^i_{X0} = 0$.
\end{proof}
% \avnote{need to check if want to keep the following corollary.}
% \begin{cor}\label{cor:opt-single}
% In a $k$-stage screening process with a single group $A$ whose tests satisfy the ``minimally effectiveness'' property and for all $i\in [k], \tau^i_{A1} <1$, the set of policies satisfying the following property strictly Pareto dominates all policies w.r.t.~precision and recall, $\forall i\in [k],\quad (1 - \pi^i_{A1}) \pi^i_{A0} = 0$.
% % \begin{align}%\label{eq:pareto-front-single}
% %     \forall i\in [k],\quad (1 - \pi^i_{A1}) \pi^i_{A0} = 0.
% % \end{align}
% \end{cor}

Next, we show additional structures of the set of Equal Opportunity policies ${\mathcal P}_{k, {\mathcal X}}$ that weakly Pareto dominates {\em all} Equal Opportunity policies.

\begin{lem}\label{lem:pi-zero}
Consider a $k$-stage screening process whose tests satisfy the ``minimal effectiveness'' property.
The set of Equal Opportunity policies $\mathcal{S} \subseteq \mathcal{P} = \{\pi \in [0,1]^{2|\mathcal{X}|k} : (1 - \pi^i_{X1}) \pi^i_{X0} = 0, \forall X\in \mathcal{X}, i\in [k]\}$ where for each group $X\in \mathcal X$, there exists at most one level $i \in [k]$ such that $0 < \pi^i_{X0} < 1$, weakly Pareto dominates all Equal Opportunity policies.

In other words, any Equal Opportunity policy $\pi$ of the screening process is weakly Pareto dominated by $\tilde{\pi} \in \mathcal{S}$ (in every policy $\tilde{\pi} \in \mathcal{S}$, for each group $X\in \mathcal X$, there exists at most one level $i$ such that $0 < \tilde{\pi}^i_{X0} < 1$). 
\end{lem}
\begin{proof}
Suppose for contradiction that there exist a group $X\in \mathcal X$ and levels $i,j$ such that $0 < \pi^i_{X0}, \pi^j_{X0} < 1$. Next, we show that we can modify $\pi$ in levels $i$ and $j$ and replace $\pi^i_{X0}, \pi^j_{X0}$ with $\tilde{\pi}^i_{X0}, \tilde{\pi}^j_{X0}$ such that 
\begin{align}
    M_{X,\tau^i, \pi^i} M_{X,\tau^j, \pi^j} &= (\tau^i_{X1} + \pi^i_{X0} (1- \tau^i_{X1})) (\tau^j_{X1} + \pi^j_{X0} (1- \tau^j_{X1})) \nonumber\\
    &= (\tau^i_{X1} + \tilde{\pi}^i_{X0} (1- \tau^i_{X1})) (\tau^j_{X1} + \tilde{\pi}^j_{X0} (1- \tau^j_{X1})) = M_{X,\tau^i, \tilde{\pi}^i} M_{X,\tau^j, \tilde{\pi}^j}, \label{eq:equal-opp-cond} \\
    N_{X,\tau^i, \pi^i} N_{X,\tau^j, \pi^j}&=(\tau^i_{X0} + \pi^i_{X0} (1- \tau^i_{X0})) (\tau^j_{X0} + \pi^j_{X0} (1- \tau^j_{X0})) \nonumber \\
    &> (\tau^i_{X0} + \tilde{\pi}^i_{X0} (1- \tau^i_{X0})) (\tau^j_{X0} + \tilde{\pi}^j_{X0} (1- \tau^j_{X0})) = N_{X,\tau^i, \tilde{\pi}^i} N_{X,\tau^j, \tilde{\pi}^j} \label{eq:prec-opt}
\end{align}
Note that Eq.~\eqref{eq:equal-opp-cond} guarantees that the new policy $\tilde{\pi}$ satisfies Equal Opportunity and has the same recall as the policy $\pi$. Moreover, Eq.~\eqref{eq:prec-opt} shows that precision of the new policy is not less than than the precision of $\pi$. 
Next, we show that in the new policy, either  $\tilde{\pi}^i_{X0} \in \{0,1\}$ or $\tilde{\pi}^j_{X0} \in \{0,1\}$. 

Without loss of generality, we can assume that the feasible range of values for $\tilde{\pi}^i_{X0}$ to satisfy Equal Opportunity is $[\pi^i_{X0} - \epsilon^i, \pi^i_{X0} + \delta^i]$ which corresponds to $[\pi^j_{X0} - \delta^j, \pi^j_{X0} + \epsilon^j]$. Both intervals are sub-intervals of $[0,1]$ and since both $\tilde{\pi}^j_{X0}, \tilde{\pi}^i_{X0}$ belong to $[0,1]$, it is straightforward to verify that $(\pi^i_{X0} - \epsilon^i) (1 - (\pi^j_{X0} + \epsilon^j)) = (1-(\pi^i_{X0} + \delta^i)) (\pi^j_{X0} - \delta^j) =0$.

Let $L = \frac{M_X}{\tau^i_{X1}\tau^j_{X1}}$ where $M_X = M_{X, \tau^i, \pi^i} M_{X, \tau^j, \pi^j} = M_{X, \tau^i, \tilde{\pi}^i} M_{X, \tau^j, \tilde{\pi}^j}$. By the ``minimally effectiveness'' property, $1< L < \frac{1}{\tau^i_{X1} \tau^j_{X1}}$. Then, satisfying Equal Opportunity is equivalent to satisfy the following constraint, $(1 + \tilde{\pi}^i_{X0}(\frac{1-\tau^i_{X1}}{\tau^i_{X1}})) (1 + \tilde{\pi}^j_{X0}(\frac{1-\tau^j_{X1}}{\tau^j_{X1}})) = L$.
% \begin{align*}
%     (1 + \tilde{\pi}^i_{X0}(\frac{1-\tau^i_{X1}}{\tau^i_{X1}})) (1 + \tilde{\pi}^j_{X0}(\frac{1-\tau^j_{X1}}{\tau^j_{X1}})) = L
% \end{align*}
Hence, it implies that %$\tilde{\pi}^j_{X0} 
%    = (\frac{L}{1 + \tilde{\pi}^i_{X0}(\frac{1-\tau^i_{X1}}{\tau^i_{X1}})} - 1)/(\frac{1-\tau^j_{X1}}{\tau^j_{X1}})
%    = (\frac{\tau^j_{X1}}{1-\tau^j_{X1}}) (\frac{L - 1 - \tilde{\pi}^i_{X0}(\frac{1-\tau^i_{X1}}{\tau^i_{X1}})}{1 + \tilde{\pi}^i_{X0}(\frac{1-\tau^i_{X1}}{\tau^i_{X1}})})$.
\begin{align*}%\label{eq:pi-j}
    \tilde{\pi}^j_{X0} 
    = (\frac{L}{1 + \tilde{\pi}^i_{X0}(\frac{1-\tau^i_{X1}}{\tau^i_{X1}})} - 1)/(\frac{1-\tau^j_{X1}}{\tau^j_{X1}})
    = (\frac{\tau^j_{X1}}{1-\tau^j_{X1}}) (\frac{L - 1 - \tilde{\pi}^i_{X0}(\frac{1-\tau^i_{X1}}{\tau^i_{X1}})}{1 + \tilde{\pi}^i_{X0}(\frac{1-\tau^i_{X1}}{\tau^i_{X1}})})
\end{align*}

\paragraph{Case 1: $\max(\tau^i_{X1}, \tau^j_{X1}) = 1$.} Without loss of generality, suppose $\tau^i_{X1} = 1$. Then, we can simply set $\tilde{\pi}^i_{X0} = 0$ and the resulting policy $\tilde{\pi}$ will maintain Equal Opportunity. Moreover, since $1-\tau^i_{X0} > 0$, $N_{X, \tau^i,\tilde{\pi}^i} \leq N_{X, \tau^i,\pi^i}$. In the other case, we can similarly set $\tilde{\pi}^j_{X0}=0$.

\paragraph{Case 2: $\tau^i_{X1}, \tau^j_{X1} < 1$.}
The task of finding $\tilde{\pi}^i_{X0}$ is as follows:
    \begin{align*}
        \tilde{\pi}^i_{X0}&=\argmin_{y\in [\pi^i_{X0} - \epsilon^i, \pi^i_{X0} + \delta^i]} f(y) := (\tau^i_{X0} + y (1- \tau^i_{X0})) (\tau^j_{X0} + (\frac{\tau^j_{X1}}{1 - \tau^j_{X1}})(\frac{L - 1 - y (\frac{1-\tau^i_{X1}}{\tau^i_{X1}})}{1 + y (\frac{1-\tau^i_{X1}}{\tau^i_{X1}})})(1-\tau^j_{X0})) 
    \end{align*}
Next, we show that for any $y \in [0,1]$, $f''(y) =- \frac{2L(\tau^j_{X0} -1)(\frac{\tau^j_{X1}}{1-\tau^j_{X1}})(\frac{1-\tau^i_{X1}}{\tau^i_{X1}}) (\frac{\tau^i_{X0}}{\tau^i_{X1}} - 1)}{(1 + (\frac{1-\tau^i_{X1}}{\tau^i_{X1}}) y)^3} < 0$.
    % \begin{align*}
    %     f''(y) =- \frac{2L(\tau^j_{X0} -1)(\frac{\tau^j_{X1}}{1-\tau^j_{X1}})(\frac{1-\tau^i_{X1}}{\tau^i_{X1}}) (\frac{\tau^i_{X0}}{\tau^i_{X1}} - 1)}{(1 + (\frac{1-\tau^i_{X1}}{\tau^i_{X1}}) y)^3} < 0.
    % \end{align*}
To prove it note that the minimal ``effectiveness property'' of the tests $\{\tau^i\}_{i\in [k]}$ (i.e., $\tau^i_{X1} > \tau^i_{X0} \ge 0, \forall X\in \mathcal X, i\in [k]$) implies that $\frac{\tau^i_{X0}}{\tau^i_{X1}} -1 < 0$. Moreover since by our assumption $\tau^j_{X1}, \tau^i_{X1} <1$, $f''(y)<0$ for all values of $y \in [0,1]$. Since $f$ is a concave function in $[\pi^i_{X0} - \epsilon^i, \pi^i_{X0} + \delta^i]$, the minimum value of $f$ in this interval obtained in one of its endpoints. In other words, the maximum precision corresponds to the case either $\tilde{\pi}^i_{X0} \in \{0,1\}$ or $\tilde{\pi}^j_{X0} \in \{0,1\}$.
\end{proof}

Finally, we show that each group can only have at most one level that partially uses its corresponding test.
\begin{lem}\label{lem:one-partial}
Consider a $k$-stage screening process whose tests satisfy the ``minimal effectiveness'' property.
The set of Equal Opportunity policies $\mathcal{P}_{k,\mathcal X} \subset \mathcal{P} = \{\pi \in [0,1]^{2|\mathcal{X}|k} : (1 - \pi^i_{X1}) \pi^i_{X0} = 0, \forall X\in \mathcal{X}, i\in [k]\}$ where for each group $X\in \mathcal X$, there exists at most one level $i \in [k]$ such that $\pi^i_{X1} < 1$ or $0<\pi^i_{X0} <1$, weakly Pareto dominates all Equal Opportunity policies.
\end{lem}
The proof is similar to the proof of Lemma~\ref{lem:pi-zero} and we defer it to Appendix~\ref{sec:linear-objective-proofs}.
The above lemma enforces a very restricted structure on the set $\mathcal{P}_{k, \mathcal X}$ of Equal Opportunity policies that {\em weakly Pareto dominate} all Equal Opportunity policies. To summarize, in each policy $\pi \in \mathcal{P}_{k, \mathcal X}$, for each group $X\in {\mathcal X}$, the restriction of $\pi$ on $X$ has the following properties
\begin{enumerate}
    \item There is at most one level $i^*\in [k]$ such that $\pi$ {\em partially uses} the test $\tau^{i^*}$; i.e., {\em either}
    $0< \pi^{i^*}_{X1} < 1$ and $\pi^{i^*}_{X0} = 0$, {\em or} $\pi^{i^*}_{X1} = 1$ and $0< \pi^{i^*}_{X0} < 1$.
    
    \item In any remaining level $i$, $\pi^i$ either {\em bypasses} $\tau^i$ (i.e., $\pi^i_{X1} \pi^i_{X0} =1$), or {\em fully exploits} $\tau^i$ (i.e., $\pi^i_{X1} = 1, \pi^i_{X0} =0$).
\end{enumerate}

\begin{thm}[\bf Exact Algorithms for Linear Combination of Precision and Recall]
Given any linear objective function of form $f_\alpha(\pi) := \alpha \cdot \mathrm{precision}(\pi) + (1-\alpha) \cdot \mathrm{recall}(\pi)$, There exists an exact algorithm that runs in time $O(k^{|{\mathcal X}|} \cdot 2^{k|{\mathcal X}|})$ and finds an Equal Opportunity policy of the screening process with parameters $(q,u, \tau, \mathcal X)$ that maximizes $f_\alpha$.
\end{thm}
\begin{proof}
Using the aforementioned set ${\mathcal P}_{k, {\mathcal X}}$ of weakly Pareto optimal policies (w.r.t. precision and recall) that satisfy the Equality of Opportunity, we enumerate over all policies in ${\mathcal P}_{k, {\mathcal X}}$ as follows.
\begin{itemize}
    \item For each group $X\in \mathcal X$, pick a level $i_X \in [k]$ ({\em i.e., $k^{|{\mathcal X}|}$ possible configurations}).
    \item Fix an ``integral'' policy $\pi$ for the rest of levels in each group $X\in \mathcal X$,
    \begin{itemize}
        \item In each group $X\in \mathcal X$, for each level $i \neq i_X$, we decide whether to fully use the test ($\pi^i_{A1} =1, \pi^i_{A0}=0$) or to bypass the test ($\pi^i_{X1} = \pi^i_{X0} = 1$) ({\em i.e., $2^{(k-1)|{\mathcal X}|}$ possible configurations}). 
    \end{itemize}
    \item For each $X\in {\mathcal X}, i_{X}\in [k]$, we fix the policy $\pi^{i_X}$ partially as follows,
    \begin{itemize}
        \item $(1 - \pi^{i_X}_{X1}) \pi^{i_X}_{X0} = 0, \forall X\in {\mathcal X}$ ({\em i.e., $2^{|{\mathcal X}|}$ possible configurations}). 
    \end{itemize}
\end{itemize}
In each of the policies $\pi$ as constructed above, we set the remaining $\pi$ values (i.e., $\pi^{i_X}$) so that Equality of Opportunity is satisfied and the objective function $f_\alpha$ is maximized.  
Finally, we maintain the configuration $\pi$ that maximizes $f_\alpha$. Note that the whole process takes $O(k^{|{\mathcal X}|} \cdot 2^{k|{\mathcal X}|})$ time.  
\end{proof}

Similarly, we can show the following.
\begin{thm}[\bf Exact Algorithms for Linear Combination of reciprocal of Precision and Recall]
Given any objective function $g_\alpha(\pi) := \alpha/\mathrm{precision}(\pi) + (1-\alpha) / \mathrm{recall}(\pi)$, There exists an exact algorithm that runs in time $O(k^{|{\mathcal X}|} \cdot 2^{k|{\mathcal X}|})$ and finds an Equal Opportunity policy of the screening process with parameters $(q,u, \tau, \mathcal X)$ that minimizes $g_\alpha$.
\end{thm}

\begin{rem}[{\bf General Objective Functions}]\label{rem:exact-alg}
Our approach provides an exact algorithm for maximizing (resp., minimizing) a given pipeline efficiency objective $f$ (resp., pipeline complexity cost $g$) over Equal Opportunity policies if $f$ (resp., $g$) satisfies the following natural condition: 
for any pair of policies $\pi_1, \pi_2$ where $\pi_1$ weakly Pareto dominates $\pi_2$ w.r.t. precision and recall, $f(\pi_1) \ge f(\pi_2)$ (resp., $g(\pi_1) \le g(\pi_2)$).   
\end{rem}

\subsection{An FPTAS Algorithm}\label{sec:fptas}
In this section, we present FPTAS algorithms for maximizing a given pipeline efficiency objective (resp., minimizing a given pipeline cost function) while satisfying the Equal Opportunity requirement.
We consider two regimes. 
In this section, as in previous sections, we consider the regime where we are allowed to treat individuals from different groups differently; more precisely, we can set $\pi^j_{Xi} \neq \pi^j_{Yi}$ for $j\in [k], i\in\{0,1\}$. Next, in Section~\ref{sec:single-policy}, we consider a new regime where the goal is to achieve Equal Opportunity while treating individuals from both groups similarly; $\forall i\in [k], X\neq Y\in {\mathcal X}, \pi^i_{X1} = \pi^i_{Y1}, \pi^i_{X0} = \pi^i_{Y0}$.

To exploit our algorithm in different settings, we describe it for the most basic setting of the problem. Given a single group of applicants with parameters $q,u$ and a pipeline $\pipeline = \{\tau^i\}_{i\in [k]}$, the goal is find a policy $\pi$ that maximizes a given {\em pipeline efficiency objective} $f(\mathrm{recall}(\pi, \pipeline), \mathrm{precision}(\pi, q, u ,\pipeline))$. Our approach works for a quite general set of objective functions; more notably, as in the previous section, for two natural settings: maximizing a {\em linear combination of precision and recall} and minimizing a {\em linear combination of reciprocals of precision and recall}. 

\paragraph{High-level Description of Algorithm.}
Now we write a dynamic program (DP) to optimize a given pipeline efficiency objective $f$ up to a given accuracy parameter $\epsilon$.
We create a DP-table $M[i, \p, \n]$ where $i\in [k]$, $\p \in [0,\ell_{\p} := \log_{1-\epsilon} L_{\p}]$ and $\n \in [0,\ell_{\n} := \log_{1-\epsilon} L_{\n}]$ where $L_{\p}, L_{\n}$ are lower bounds on {\em True Positive Rate} and {\em False Positive Rate} respectively. 
For each set of parameters $(i, \p, \n)$, $M[i,\p,\n]$  will be a Boolean value indicating whether there exists a policy such that by the end of level $i$, the true positive rate becomes at least $(1-\epsilon)^\p$ and the False Positive Rate becomes at most $(1-\epsilon)^\n$. 
Without loss of generality and for the simplicity of the exposition, we assume $L_{\p}$ and $L_{\n}$ are powers of $(1-\epsilon)$; otherwise we can simply round the lower bounds to largest powers of $(1-\epsilon)$ smaller than actual bounds. 

\paragraph{Solving the DP} We fill out the DP table starting from $i=1$ as follows. First, for any $j_0 \in [0,\ell_{\n}], j_1 \in [0, \ell_{\p}]$, $M[1, j_1, j_0] = \mathrm{true}$ iff the following system of linear inequalities has a feasible solution.
\begin{align} 
    \tau^1_0 x + (1- \tau^1_0) y &\leq (1-\epsilon)^{j_0}, &\tau^1_1 x + (1- \tau^1_1) y \geq (1-\epsilon)^{j_1} &&\label{eq:base-rule}
\end{align}

Next, we describe the update rule for $i>1$. For any $\p \in [0, \ell_{\p}], \n \in [0, \ell_{\n}]$, $M[i+1, \p, \n] = \bigvee_{(j_0, j_1) \in {\mathcal F}_{i+1}} M[i, {\p}-{j_1}, {\n}-{j_0}]$,
% \begin{align*}
%     M[i+1, \p, \n] = \bigvee_{(j_0, j_1) \in {\mathcal F}_{i+1}} M[i, {\p}-{j_1}, {\n}-{j_0}], 
% \end{align*} 
where ${\mathcal F}_{i+1}$ is a set of $(j_0 \le \n, j_1 \le \p)$ for which the following linear program has a feasible solution,
\begin{align}
    \tau^{i+1}_0 x+ (1-\tau^{i+1}_0) y &\le (1-\epsilon)^{j_0},
    &\tau^{i+1}_1 x+ (1-\tau^{i+1}_1) y \ge (1-\epsilon)^{j_1}. &&\label{eq:update-rule}
\end{align}

Note that $x, y$ can be interpreted as $\pi^{i+1}_1, \pi^{i+1}_0$, respectively. Moreover, the system of linear inequalities of the update rule in level $i+1$ (Eq.~\eqref{eq:update-rule}) is similar to the rules for the base case (Eq.~\eqref{eq:base-rule}). 
% Refer to Figure~\ref{fig:system-linear-inequalities}. 
% \begin{figure}[!h]
% \begin{center}
% \includegraphics[scale=.5]{dp_fig}
% \caption{The gray region, which is the set of points $(x,y)$ below the red line, above the blue line and inside $\{0\le x,y \le 1\}$, denotes the feasible region for the system of linear inequalities with $(j_0, j_1, \tau_0, \tau_1, \epsilon)$ as in the left plot. The system of linear inequalities with $(i_0, i_1, \tau_0, \tau_1, \epsilon)$ as in the right plot has no feasible solution.}\label{fig:system-linear-inequalities}
% \end{center}
% \end{figure}
% \avnote{maybe remove the figure?}
\begin{lem}\label{lem:DP}
For any $i\in [k]$, if there exists a policy $\pi$ with True Positive Rate $\t_i \ge L_{\p}/(1-\epsilon)^{i-1}$ and False Positive Rate $\f_i$ by the end of level $i$, then for any $j_1\in [0, \ell_{\p}], j_0 \in [0, \ell_{\n}]$ with $(1-\epsilon)^{j_1} \ge \t_i \cdot (1-\epsilon)^{i-1}$ and $(1-\epsilon)^{j_0} \le \min\{ 1, \max\{ L_{\n}, \f_i\}/ (1-\epsilon)^{i-1}\}$, $M[i, {j_1}, {j_0}] = \mathrm{true}$.

In other words, if the policy $\pi$ exists then the DP approach finds a policy with true positive rate at least $(1-\epsilon)^{j_1}$ and false positive rate at most $(1-\epsilon)^{j_0}$.
\end{lem}
The proof is deferred to Section~\ref{sec:linear-objective-proofs}.

\begin{lem}[\bf DP Main Lemma]\label{lem:DP-table-runtime}
For any group $X\in {\mathcal X}$, an accuracy parameter $\epsilon$ and lower bounds on the false positive rate, $L_{\n}$, and the true positive rate, $L_{\p}$, if there exists a policy $\pi^*$ with true positive rate $\t \ge L_{\p}/(1-\epsilon)^{k-1}$ and false positive rate $\f>0$, then the DP algorithm runs in time $O(\frac{k \log^2 (1/L_{\p}) \log^2 (1/L_{\n})}{\epsilon^4})$ and finds a policy $\pi$ with true positive rate at least $(1-\epsilon)^{k-1} \cdot \t$ and false positive rate at most $\min\{ 1, \max\{ L_{\n}, \f\}/ (1-\epsilon)^{k-1}\}$.
\end{lem}
\begin{proof}
The size of table is $O(k \ell_{\p} \ell_{\n})$ and updating each entry in the table takes $O(\ell_\p \ell_\n)$. Hence, the total runtime to compute all entries in the DP table is $O(k\ell^2_\p \ell^2_\n) = O(\frac{k \log^2(1/L_{\p}) \log^2(1/L_{\n})}{\epsilon^{4}})$.

Now we apply the DP approach and by Lemma~\ref{lem:DP}, the solution returned by the algorithm has the true positive rate and the false positive rate satisfying the guarantee of the statement.
\end{proof}

\paragraph{Implications of DP} Here we present FPTAS algorithms using the described DP approach in different settings. We state the results formally and their proofs are deferred to Appendix~\ref{sec:linear-objective-proofs}.

\begin{thm}[\bf FPTAS for Linear Combination of Precision and Recall]\label{thm:EQ-DP-max}
Consider a $k$-stage screening process with parameters $(u, q, \tau, \mathcal X)$ and for any policy $\pi$, let $f_{\alpha}(\pi) = (1-\alpha) \cdot \mathrm{recall}(\pi) + \alpha \cdot \mathrm{precision}(\pi)$ where $\alpha > 0$. Given an accuracy parameter $\epsilon$, there exists an FPTAS that runs in time $O(\frac{|{\mathcal X}| k^5 \log^4(1/\epsilon)}{\epsilon^4})$ and finds an Equal Opportunity policy $\pi$ such that $f_{\alpha}(\pi) \ge (1-\epsilon)f_{\alpha}(\pi^*)$ where $\pi^*$ maximizes $f_\alpha$ over Equal Opportunity policies. 
\end{thm}

\begin{thm}[\bf FPTAS for Linear Combination of Reciprocals Precision and Recall]\label{thm:EQ-DP-min}
Consider a $k$-stage screening process with parameters $(u, q, \tau, \mathcal X)$ and for any policy $\pi$, let $g_{\alpha}(\pi) = (1-\alpha)/\mathrm{recall}(\pi) + \alpha / \mathrm{precision}(\pi)$ where $\alpha > 0$. Given an accuracy parameter $\epsilon$, there exists an FPTAS that runs in time $O(\frac{|{\mathcal X}| k^7 (\log^2(1/\epsilon) + k^2)}{\epsilon^4})$ and finds an Equal Opportunity policy $\pi$ such that $g_{\alpha}(\pi) \le (1+\epsilon)g_{\alpha}(\pi^*)$ where $\pi^*$ minimizes $g_\alpha$ over Equal Opportunity policies. 
\end{thm}
% The proof of the above theorem is similar to the proof of Theorem~\ref{thm:EQ-DP-max} and is deferred to Section~\ref{sec:linear-objective-proofs}.
%\paragraph{General Function of Precision and Recall.} 
\begin{rem}[{\bf General Objective Functions}]\label{rem:fptas}
In Theorem~\ref{thm:EQ-DP-max} and~\ref{thm:EQ-DP-min} we presented FPTAS for finding Equal Opportunity policies optimizing two standard pipeline efficiency objective functions. 
Here, we generalize the above theorems when the pipeline efficiency objective function $f: [0,1]^{2} \rightarrow \mathbb{R}$ which maps precision and recall to efficiency scores have certain properties. 
Also, we define $g:[0,1]^{2} \rightarrow \mathbb{R}$ such that for any $\t,\f \in [0,1]^{2}$, $g(\t, \f) := f(\mathrm{recall}(\t),\mathrm{precision}(\t,\f))$.
We describe the properties when the goal is to maximize $f$---{\em the required conditions for the minimization version is similar}.
\begin{itemize}
    \item $f$ is non-decreasing w.r.t.~both precision and recall---{\em equivalently, $g$ is non-decreasing in $\t$ and non-increasing in $\f$}.
    \item There exist $L_{\p}, L_{\n} >0$ such that there exists a $(1-\alpha)$-approximate solution of $f$ with $\t \in (L_{\p},1], \f \in (L_{\n},1]$.
    \item The function $f$ is $\beta$-Lipschitz on $\{(x,y) | x\in (L_{\p}, 1], y \in (L_{\n},1]\}$.
\end{itemize}
In particular, the above properties are sufficient to show that the DP approach finds a $(1-\epsilon)$-approximation of $f$ in time $\mathrm{poly}(k, |\mathcal{X}|, \epsilon^{-1}, \log(1/L_{\p}), \log(1/L_{\n}))$.
\end{rem}

%\subsection{Selecting from Available Tests}
\begin{rem}[{\bf Selecting from Available Tests}]\label{rem:select-tests}
Suppose that in contrast to our previous approaches, we do allow for the design of the pipeline in that we allow the firm to select some tests to create a pipeline.
For instance, imagine that there is a budget and the firm is allocating this budget to buy tests.
The goal of the firm is the same, e.g. to exhibit a pipeline satisfying a fairness requirement. 
Our algorithms can be modified to handle to this case by adding a term in the DP table corresponding to the budget remaining, with a decision point of choosing to use a given test or not.  Note that the ordering of tests in the pipeline does not matter for the objectives considered.
\end{rem}

\section{Alternate Models}\label{sec-altmodels} In this section we describe some alternate settings, such as using a single promotion policy for both demographic groups (which might be required by regulation), or requiring Equalized Odds.
%, and requiring the fairness criterion hold independently at every step of the screening process. 
%These alternate scenarios evince some of the relative strengths of our primary scenario, Equal Opportunity with two screening processes.
%However, these alternate scenarios are of practical relevance and are worth discussing independently.

\subsection{Screening Processes with Same Policy for All Groups} \label{sec:single-policy}
One alternate fairness model is to additionally require the same policy be used for all groups.
%there are \textit{not} multiple classifiers fine-tuned for each group. 
While utilizing demographic features can aid in achieving fairness goals (e.g.~\cite{dwork2012fairness, hardt2016equality}), 
%in alternative words, this could mean giving two individuals of different demographic groups  different predictions or decisions even if 
%every other feature in their corresponding data point is identical. 
in some regulatory regimes, this fairness-through-awareness may be illegal or problematic, even when intended to ensure equitable treatment. 

In our setting,  if we are constrained to follow \textit{group-blindness}, there be would only one set of tests and one ordering of the tests that all applicants are tested on. Analogously to the previous setting, the action space of the algorithm remains modifying the promotion probabilities, but we now only have one set of policies to modify. 
We also exhibit a DP algorithm for this setting, which we defer to Section~\ref{sec:single-policy-app}.
However, a simple example shows the inefficiencies in this regime. 
Suppose we have a single test with $T_{A} = (1,0)$ and $T_B =(1/2,0).$
Observe that since we are constrained to use group blindness and satisfy Equal Opportunity, there is no way to use the test without violating Equal Opportunity.  Thus, the only way to satisfy Equal Opportunity is to completely bypass the test.

\subsection{Equalized Odds \label{eodds}}
Next, recall that the requirement of Equalized Odds mandates equal True Positive and False Positive rates for all groups.
%, a stronger requirement than simply requiring Equal Opportunity. 
%Now we consider the analogous results and algorithms for satisfying Equalized Odds in our screening process. 
In the appendix, we show structural properties of an optimal promotion policy that satisfies Equalized Odds. 
However, we also note the interview efficiency cost (precision) of requiring Equalized Odds.
In particular, the gap between the interview efficiency of $\pi_{\mathrm{EOdd}}$ and $\pi_{\mathrm{EOpp}}$ can be as large as $\frac{1}{q}-\epsilon$ for any arbitrary $\epsilon >0$.
See Theorem \ref{thm:Eodd-vs-Eopp} for details. 

\subsection{Discussion  Comparing Equalized Odds and Equal Opportunity} 
From the perspective of a decision maker in the wild, how to interpret and operationalize these results?
A robust take-away is that requiring Equalized Odds and Equal Opportunity have substantially different efficiency consequences.
Based on our examples, it seems unlikely that Equalized Odds is effective in this model, especially when requiring Equalized Odds at each stage.
In contrast, the fact that requiring Equal Opportunity at each stage is equivalent to requiring Equal Opportunity of the overall process with respect to interview efficiency may have benefits in ensuring public confidence in the model. 

\subsection{Intersectionality \label{intersect}}
A natural question is how to think when the demographic groups may have an arbitrarily overlapping structure. 
This suggests several open questions in our model, e.g. if a person is in groups $A$ and $B$, then which test parameter $\tau_{A}$ or $\tau_{B}$ corresponds to that person?
Perhaps a direction is to assign to that person an interpolation between these values.
A naive approach is when there are $k$ groups, to create $2^{k}$ new groups and $2^{k}$ test parameters corresponding to every possible group intersection.
If $k$ is small, this may be computationally feasible, but is not responsive when the relevant sub-groups/intersections may not be known apriori.
Perhaps our model could be merged with multi-calibration notions~\citep{pmlr-v80-hebert-johnson18a}.
%Extensions to these models would include more expressive labels e.g. a real valued score for labels with high scores being preferred or the tests returning scores as well, rather than binary outcomes in our model. 

\section{Conclusions}
In contrast to some fairness in machine learning work, we focus on post-processing fairness modifications, rather than thinking about the fairness problem in screening processes where tests can be designed from scratch. 
While we believe that the more a priori design approach will have substantial benefits in practice, our approach of modifying pre-existing tests, combined with a concrete (and simple to evaluate) fairness notion, Equal Opportunity, is closely aligned with real world circumstances and models, especially in short term and iterative improvements to models. In some settings, the firm making hiring decisions will outsource some aspects of its pipeline to third party companies and the tests will be a black box, but possibly that come with statistics that can be used in our algorithms.  
This decoupling allows the effective implementation of fairness aware promotion policies in the short term. 

%% file: missing-proofs.tex
\section{Proofs from Section~\ref{maxprecsection}}\label{sec:precision-max-proof}
\begin{proof}[Proof of Theorem~\ref{thm:multi-eq-opp-prec}]
First, we show that for any $M\in (0,1]$, any Equal Opportunity policy $\pi_M$ with recall $M$ has interview efficiency at most
\begin{align}
\ie(q, u, \tau, \pi_M) 
= \frac{\sum_{X\in \mathcal X}\q_X M_{X, \tau, \pi_M}}{\sum_{X\in \mathcal X}\q_X M_{X, \tau, \pi_M} + \u_X N_{X, \tau, \pi_M}} &= %\frac{q M}{q M + \sum_{X\in \mathcal X} \u_X N^X_{\tau, \pi_M}} = 
\frac{\|q\|_1}{\|q\|_1 + \sum_{X\in \mathcal X} \u_X \frac{N_{X, \tau, \pi_M}}{M}} \nonumber \\
&\leq \frac{\|q\|_1}{\|q\|_1 + \sum_{X\in \mathcal X} \u_X \Pi_{i=1}\frac{\tau^i_{X0}}{\tau^i_{X1}}}, \label{eq:upper-bound-interview-eff-mult}  
\end{align}
where the last inequality follows from the minimally effectiveness of tests in the screening process and an argument identical to Eq.~\eqref{eq:X-bound}. Note that the inequality holds no matter what the value of $M$ is.
Next, we show that opportunity ratio policy achieves the maximum possible interview efficiency as shown in Eq.~\eqref{eq:upper-bound-interview-eff-mult}. Let $X^* = \argmin_{X\in \mathcal X} \Pi_{j\in [k]}\tau^j_{X1}$. Recall that the opportunity ratio policy $\pi$ is defined as follow.
\begin{align*}
    \pi^1_{X0} &= 0 \text{ and } \pi^1_{X1} = \Pi_{i\in [k]}(\tau^i_{X^*1}/\tau^i_{X1}) &&\forall X\in \mathcal X \\
    \pi^i_{X0} &= 0 \text{ and } \pi^i_{X1} = 1 &&\forall X\in {\mathcal X}, i\ge 2
\end{align*}
It is straightforward to check that $\pi$ is an Equal Opportunity policy with recall $\Pi_{i\in [k]} \tau^i_{X^*1}$. Moreover, the interview efficiency of $\pi$ is 
\begin{align*}
\ie(q, u, \tau, \pi)     
    = \frac{\sum_{X\in \mathcal X}\q_X M_{X, \tau, \pi}}{\sum_{X\in \mathcal X}\q_X M_{X, \tau, \pi} + \u_X N^X_{\tau, \pi}} %\\
    &= \frac{\sum_{x\in \mathcal{X}}q_X \Pi_{i\in [k]} \tau^i_{X^*1}}{\sum_{x\in \mathcal{X}}q_X \Pi_{i\in [k]} \tau^i_{X^*1} + \sum_{X\in \mathcal X}\u_X \Pi_{i\in [k]} \frac{\tau^i_{X^*1}\tau^i_{X0}}{\tau^i_{X1}}} \\
    &= \frac{\|q\|_1}{\|q\|_1 + \sum_{X\in \mathcal X}\u_X \Pi_{i=1}^k\frac{\tau^i_{X0}}{\tau^i_{X1}}}
\end{align*}
Hence, $\pi$ is the Equal Opportunity policy maximizing the interview efficiency.
\end{proof}

\section{Proofs from Section~\ref{sec:linear-objective}}\label{sec:linear-objective-proofs}

\begin{proof}[Proof of Claim~\ref{clm:non-zero-pi-one}]
Suppose for contradiction that there exists a group $X\in \mathcal X$ and a level $i\in [k]$ such that $\pi^i_{X1} = 0$. 
First note that $(1-\tau^i_{X1}) \pi^i_{X0} >0$; otherwise, the policy is useless because it prevents candidates of group $X$, in particular the qualified ones, from reaching the interview stage. Hence, by the Equal Opportunity requirement, no qualified candidate will reach the interview stage. 

Next, we show that there exists a policy $\tilde{\pi}$ (which only differs from $\pi$ in level $i$ of group $X$) that satisfies Equal Opportunity for the given screening process and strictly Pareto dominates $\pi$; $\tilde{\pi}^i_{X1}= (\frac{1-\tau^i_{X1}}{\tau^i_{X1}}) \pi^i_{X0}$ and $\tilde{\pi}^i_{X0} = 0$.
% \begin{align*}
%     \Big(\tilde{\pi}^i_{X1}&= (\frac{1-\tau^i_{X1}}{\tau^i_{X1}}) \pi^i_{X0},\tilde{\pi}^i_{X0} = 0\Big) &&\Big(\tilde{\pi}^j_{Y1} = \pi^j_{Y1}, \tilde{\pi}^j_{Y0} = \pi^j_{Y0}\Big) \quad \text{if $(Y\neq X) \vee (j\neq i)$}
% \end{align*}

Since $M_{X, \tau^i, \tilde{\pi}^i} = \tau^i_{X1} \tilde{\pi}^i_{X1} + (1-\tau^i_{X1}) \tilde{\pi}^i_{X0} = \tau^i_{X1} \tilde{\pi}^i_{X1} = (1-\tau^i_{X1}) \pi^i_{X0} = \tau^i_{X1}\pi^i_{X1} + (1-\tau^i_{X1}) \pi^i_{X0} = M_{X, \tau^i, \pi^i}$ and $\pi$ satisfies the Equal Opportunity, $\tilde{\pi}$ also satisfies Equal Opportunity and has the same recall as $\pi$. Moreover, since
$N_{X, \tau^i, \tilde{\pi}^i} = \tilde{\pi}^i_{X1} \tau^i_{X0} = (\frac{1-\tau^i_{X1}}{\tau^i_{X1}})\pi^i_{X0} \tau^i_{X0} < (1-\tau^i_{X0}) \pi^i_{X0} = N_{X, \tau^i, \pi^i}$, $\mathrm{precision}(\tilde{\pi}) > \mathrm{precision}(\pi)$. Note that $\frac{1- \tau^i_{X1}}{\tau^i_{X1}} < \frac{1- \tau^i_{X0}}{\tau^i_{X0}}$ holds by the minimal effectiveness property of tests.
%if for all $i\in [k]$, $\tau^i_{X1} < 1$, by the minimal effectiveness property, $\mathrm{precision}(\tilde{\pi}) > \mathrm{precision}(\pi)$.
\end{proof}

%\begin{align*}
 %   & (\frac{1-\tau_1}{\tau_1} )\pi_0 \tau_0 \leq (1-\tau_0) \pi_0 \\
  %  & (1-\tau_1) \tau_0 \leq (1-\tau_0) \tau_1 \\
  %  & \tau_0 - \tau_1 \tau_0 \leq \tau_1 - \tau_1 \tau_0 \\
   % & \tau_0 \leq \tau_1 \quad \text{(minimal effectiveness)}
%\end{align*}

\begin{lem}\label{lem:pi-one}
Consider a $k$-stage screening process whose tests satisfy the ``minimal effectiveness'' property.
The set of Equal Opportunity policies $\mathcal{S} \subseteq \mathcal{P} = \{\pi \in [0,1]^{2|\mathcal{X}|k} | (1 - \pi^i_{X1}) \pi^i_{X0} = 0, \forall X\in \mathcal{X}, i\in [k]\}$, where for each group $X\in \mathcal X$, there exists at most one level $i \in [k]$ such that $\pi^i_{X1} < 1$, weakly Pareto dominates all Equal Opportunity policies.
\end{lem}
\begin{proof}
Suppose for contradiction that there are two levels $i,j$ such that $\pi^i_{X1}, \pi^j_{X1} < 1$. First note that by Claim~\ref{clm:non-zero-pi-one}, $\pi^i_{X1}, \pi^j_{X1} > 0$. Moreover, by Lemma~\ref{lem:pi-cond}, since $\pi^i_{X1}, \pi^j_{X1} < 1$, $\pi^i_{X0} = \pi^j_{X0} = 0$.

Next, we show that we can modify $\pi$ in levels $i$ and $j$ and replace $\pi^i_{X0}, \pi^j_{X0}$ with $\tilde{\pi}^i_{X0}, \tilde{\pi}^j_{X0}$ as follows: $\tilde{\pi}^i_{X1} = \pi^i_{X1} \pi^j_{X1}$ and $\tilde{\pi}^j_{X1} = 1$. Then, $M_{X, \tau^i, \tilde{\pi}^i} M_{X, \tau^j, \tilde{\pi}^j}
    = (\tilde{\pi}^i_{X1} \tau^i_{X1}) (\tilde{\pi}^j_{X1} \tau^j_{X1}) 
    = (\pi^i_{X1} \tau^i_{X1}) (\pi^j_{X1} \tau^j_{X1}) 
    = M_{X, \tau^i, \pi^i} M_{X, \tau^j, \pi^j}$.
% \begin{align*}
%     M^X_{\tau^i, \tilde{\pi}^i} M^X_{\tau^j, \tilde{\pi}^j}
%     = (\tilde{\pi}^i_{X1} \tau^i_{X1}) (\tilde{\pi}^j_{X1} \tau^j_{X1}) 
%     = (\pi^i_{X1} \tau^i_{X1}) (\pi^j_{X1} \tau^j_{X1}) 
%     = M^X_{\tau^i, \pi^i} M^X_{\tau^j, \pi^j}.
% \end{align*}
In other words, the policy $\tilde{\pi}$ satisfies Equal Opportunity and has the same recall as $\pi$. Similarly, this modification does not decrease precision. Formally, $N_{X, \tau^i, \tilde{\pi}^i} N_{X, \tau^j, \tilde{\pi}^j}
    = (\tilde{\pi}^i_{X1} \tau^i_{X0}) (\tilde{\pi}^j_{X1} \tau^j_{X0}) 
    = (\pi^i_{X1} \tau^i_{X0}) (\pi^j_{X1} \tau^j_{X0}) 
    = N_{X, \tau^i, \pi^i} N_{X, \tau^j, \pi^j}$.
% \begin{align*}
%     N^X_{\tau^i, \tilde{\pi}^i} N^X_{\tau^j, \tilde{\pi}^j}
%     = (\tilde{\pi}^i_{X1} \tau^i_{X0}) (\tilde{\pi}^j_{X1} \tau^j_{X0}) 
%     = (\pi^i_{X1} \tau^i_{X0}) (\pi^j_{X1} \tau^j_{X0}) 
%     = N^X_{\tau^i, \pi^i} N^X_{\tau^j, \pi^j}.
% \end{align*}
Hence, for each policy $\pi$, there exists another policy with at most one level $i\in [k]$ such that $\pi^i_{X0} <1$ and weakly Pareto dominates $\pi$.  
\end{proof}

\begin{proof}[Proof of Lemma~\ref{lem:one-partial}]
We follow a similar arguments as in the proof of Lemma~\ref{lem:pi-zero}. Note that by Lemma~\ref{lem:pi-zero} and Lemma~\ref{lem:pi-one} there is at most one level $i_1\in [k]$ such that $0< \pi^{i_1}_{X1} < 1$ and $\pi^{i_1}_{X0} = 0$, and there is at most one level $i_0\in [k]$ such that $\pi^{i_0}_{X1} = 1$ and $0< \pi^{i_0}_{X0} < 1$. Next, we show that we can modify the policy $\pi$ in levels $i_0$ and $i_1$ and replace $\pi^{i_0}_{X0}, \pi^{i_1}_{X1}$ with $\tilde{\pi}^{i_0}_{X0}, \tilde{\pi}^{i_1}_{X1}$ such that
\begin{align*}
    &M_{X, \tau^{i_0}, \pi^{i_0}} M_{X, \tau^{i_1}, \pi^{i_1}} 
    = (\tau^{i_0}_{X1} + \pi^{i_0}_{X0} (1- \tau^{i_0}_{X1})) (\pi^{i_1}_{X1}\tau^{i_1}_{X1}) 
    = (\tau^{i_0}_{X1} + \tilde{\pi}^{i_0}_{X0} (1- \tau^{i_0}_{X1})) (\tilde{\pi}^{i_1}_{X1}\tau^{i_1}_{X1}) 
    = M_{X, \tau^{i_0}, \tilde{\pi}^{i_0}} M_{X, \tau^{i_1}, \tilde{\pi}^{i_1}},  \\
    &N_{X, \tau^{i_0}, \pi^{i_0}} N_{X, \tau^{i_1}, \pi^{i_1}} 
    = (\tau^{i_0}_{X0} + \pi^{i_0}_{X0} (1- \tau^{i_0}_{X0})) (\pi^{i_1}_{X1}\tau^{i_1}_{X0}) 
    < (\tau^{i_0}_{X0} + \tilde{\pi}^{i_0}_{X0} (1- \tau^{i_0}_{X0})) (\tilde{\pi}^{i_1}_{X1}\tau^{i_1}_{X0}) 
    = N_{X, \tau^{i_0}, \tilde{\pi}^{i_0}} N_{X, \tau^{i_1}, \tilde{\pi}^{i_1}} 
\end{align*}
Now, we show that in the new solution, either $\tilde{\pi}^{i_0}_{X0} \in \{0,1\}$ or $\tilde{\pi}^{i_1}_{X1} = 1$. 

Without loss of generality, we can assume that the feasible range of values for $\tilde{\pi}^{i_0}_{X0}$ to satisfy Equal Opportunity is $[\pi^{i_0}_{X0} - \epsilon^{i_0}, \pi^{i_0}_{X0} + \delta^{i_0}]$ which corresponds to $[\pi^{i_1}_{X0} - \delta^{i_1}, \pi^{i_1}_{X0} + \epsilon^{i_1}]$. 
Both intervals are sub-intervals of $[0,1]$ and it is straightforward to verify that $(\pi^{i_0}_{X0} - \epsilon^{i_0}) (1 - (\pi^{i_1}_{X0} + \epsilon^{i_1})) = (1-(\pi^{i_1}_{X0} + \delta^{i_1}))=0$.

Let $L = M_X/ (\tau^{i_0}_{X1}\tau^{i_1}_{X1})$ where $M_X = M_{X, \tau^{i_0}, \pi^{i_0}} M_{X, \tau^{i_1}, \pi^{i_1}} = M_{X, \tau^{i_0}, \tilde{\pi}^{i_0}} M_{X, \tau^{i_1}, \tilde{\pi}^{i_1}}$. By the ``minimally effectiveness'' property, $0< L < 3$. Then, satisfying Equal Opportunity is equivalent to satisfy $(1 + \tilde{\pi}^{i_0}_{X0}(\frac{1-\tau^{i_0}_{X1}}{\tau^{i_0}_{X1}})) \tilde{\pi}^{i_1}_{X1} = L$, which implies that $\tilde{\pi}^{i_1}_{X1} = L/(1 + \tilde{\pi}^{i_0}_{X0}(\frac{1-\tau^{i_0}_{X1}}{\tau^{i_0}_{X1}}))$.
The task of finding $\tilde{\pi}^i_{X0}$ is as follows: 
\[\tilde{\pi}^i_{X0}=\argmin_{y\in [\pi^{i_0}_{X0} - \epsilon^{i_0}, \pi^{i_0}_{X0} + \delta^{i_0}]} f(y) := (\tau^{i_0}_{X0} + y (1- \tau^{i_0}_{X0})) (\tau^{i_1}_{X0} \cdot \frac{L}{1 + y(\frac{1-\tau^{i_0}_{X1}}{\tau^{i_0}_{X1}})}).\]
Next, we show that for any $y \in [0,1]$,
$f''(y) =\frac{2L\tau^{i_1}_{X0} (\frac{1-\tau^{i_0}_{X1}}{\tau^{i_0}_{X1}})(\tau^{i_0}_{X0} (\frac{1-\tau^{i_0}_{X1}}{\tau^{i_0}_{X1}}) + \tau^{i_0}_{X0} - 1)}{(1 + \tau^{i_0}_{X0} y)^3} < 0$.
To prove it note that the minimal ``effectiveness property'' of the tests $\{\tau^i\}_{i\in [k]}$ (i.e., $\tau^i_{X1} > \tau^i_{X0} \ge 0, \forall X\in {\mathcal X}, i\in [k]$) implies that $\frac{\tau^{i_0}_{X0}}{\tau^{i_0}_{X1}} -1 < 0$. Since $f$ is a concave function in $[\pi^{i_0}_{X0} - \epsilon^{i_0}, \pi^{i_0}_{X0} + \delta^{i_0}]$, the minimum value of $f$ in this interval obtained in one of its endpoints. In other words, the maximum precision corresponds to the case either $\tilde{\pi}^{i_0}_{X0} \in \{0,1\}$ or $\tilde{\pi}^{i_1}_{X1} = 1$.
\end{proof}

\begin{proof}[Proof of Lemma~\ref{lem:DP}]
The proof is by induction. For the base case ($i=1$), let $\t_1$ and $\f_1$ denote the true positive rate and the false positive rate of $\pi$ by the end of level $1$. 
The existence of $\pi$ guarantees that the system of inequalities Eq.~\eqref{eq:base-rule} with $(j_0 = \lfloor \log_{1-\epsilon} \f_1 \rfloor, j_1 = \lceil \log_{1-\epsilon}\t_1 \rceil \leq \ell_{\p})$ has a feasible solution.  
More precisely, by setting $(x = \pi_1, y = \pi_0)$, 
\begin{align*} 
    \tau^1_0 x + (1- \tau^1_0) y = \f_1 \le (1-\epsilon)^{\lfloor \log_{1-\epsilon} \f_1 \rfloor} = (1-\epsilon)^{j_0}, \quad \tau^1_1 x + (1- \tau^1_1) y = \t_1 \geq (1-\epsilon)^{\lceil \log_{1-\epsilon} \t_1 \rceil} = (1-\epsilon)^{j_1}
\end{align*}
Next, we consider $i>1$ and we assume that the claim holds for all values $i' < i$. Let $M_i := \tau^i_1\pi^i_1 + (1 - \tau^i_1) \pi^i_0$ and $N_i := \tau^i_0\pi^i_1 + (1-\tau^i_0)\pi^i_0$. Note that $\t_i = \t_{i-1} \cdot M_i$ and $\f_i = \f_{i-1} \cdot N_i$.

By the induction hypothesis and considering the first $i-1$ levels in the pipeline, since $\t_{i-1} \ge \t_i \ge L_{\p}/(1-\epsilon)^{i-1} > L_{\p}/(1-\epsilon)^{i-2}$ and $\f_{i-1} \ge \f_i$, there exist $j'_1\in[0, L_{\p}]$ and $j'_0\in [0, L_{\n}]\cup \{\infty\}$ such that $M[i-1, j'_1, j'_0] = \mathrm{true}$ and $(1-\epsilon)^{j'_1} \ge \t_{i-1} \cdot (1-\epsilon)^{i-2}$ and $(1-\epsilon)^{j'_0} \le \min\{1, \max\{L_{\n}, \f_{i-1}\}/(1-\epsilon)^{i-2}\}$. 
More precisely, the algorithm finds a policy $\bar{\pi}$ with true positive rate at least $(1-\epsilon)^{j'_1}$ and false positive rate at most $(1-\epsilon)^{j'_0}$.

Next, by setting $(\bar{\pi}^i_1 = \pi^i_1, \bar{\pi}^i_0 = \pi^i_0)$ and $(j_1 :=\argmin_{j} \{(1-\epsilon)^j \le \t_i(\bar\pi)\}, j_0 := \argmax_{j} \{(1-\epsilon)^j \ge \f_i(\bar\pi)\})$,  
\begin{align*}
    (1-\epsilon)^{j_1} > (1-\epsilon)\cdot \t_{i}(\bar \pi)
    &= (1-\epsilon)\cdot \t_{i-1}(\bar \pi) \cdot M_i &&\rhd\text{by definition of $j_1$}\\
    &\ge (1-\epsilon)\cdot (1-\epsilon)^{j'_1} \cdot M_i &&\rhd\text{by $\t_{i-1}(\bar\pi) \ge (1-\epsilon)^{j'_1}$}\\ 
    &\ge \t_{i-1} \cdot (1-\epsilon)^{i-1} \cdot M_i &&\rhd\text{by induction hypothesis}\\
    &= \t_i \cdot (1-\epsilon)^{i-1}. 
\end{align*}
Similarly,
\begin{align*}
    (1-\epsilon)^{j_0} < \min\{1, \frac{\f_{i}(\bar \pi)}{1-\epsilon}\} 
    &= \min\{1, N_i \cdot \frac{\f_{i-1}(\bar \pi)}{1-\epsilon}\} &&\rhd\text{by definition of $j_0$}\\
    &\le \min\{1, (1-\epsilon)^{j'_0} \cdot \frac{N_i}{1-\epsilon}\} &&\rhd\text{by $\f_{i-1}(\bar\pi) \le (1-\epsilon)^{j'_0}$} \\
    &\le \min\{1, \frac{\max\{L_{\n}, \f_{i-1}\}}{(1-\epsilon)^{i-2}} \cdot \frac{N_i}{1-\epsilon}\} &&\rhd\text{by induction hypothesis}\\ 
    &\le \min\{1, \frac{\max\{L_{\n}, \f_i\}}{(1-\epsilon)^{i-1}}\}
\end{align*}
which completes the proof.
\end{proof}

\begin{proof}[Proof of Theorem~\ref{thm:EQ-DP-max}]
First, as we are aiming for a $(1-\epsilon)$-approximation, we only need to consider $\alpha\in (\epsilon, 1-\epsilon)$. Otherwise, either the policy maximizing recall (i.e. bypassing all tests) or the policy maximizing precision (Opportunity Ratio policy) is a $(1-\epsilon)$-approximation for $f_\alpha$.  

Next we show in order to guarantee $(1-\epsilon)$-approximations of recall and precision of the policy maximizing $f_\alpha$, it suffices to run the described DP and consider estimates of $\t$ (true positive rate) and $\f$ (false positive rate) of form $(1-\bar\epsilon)^i$ for $i\in \mathbb{N}$ in intervals $[L_{\p}, 1]$ and $[L_{\n}, 1]$ respectively, where $\bar\epsilon \le \epsilon/(2k)$. We provide tight bounds for $L_{\p}$ and $L_{\n}$. 
Note that since for any policy $\pi$, the true positive rate ($\t_i$) and the false positive rate ($\f_i$) are non-decreasing in $i$, it suffices to provide ``large enough'' lowerbounds $L_{\p}$ and $L_{\n}$ for $\t$ and $\f$ in the final stage respectively. 

\paragraph{Bounding $L_{\p}$.} Consider the policy $\pi_{\bypass}$, which bypasses all the tests in both groups, i.e., $\pi^i_{X0}=\pi^i_{X1} =1$ for all $i\in [k], X\in {\mathcal X}$. Since $\pi_\bypass$ is an Equal Opportunity policy for the pipeline and $f_\alpha(\pi_\bypass) = (1-\alpha) + \alpha \|q\|_1$, any optimal Equal Opportunity policy $\pi^*$ for $f_\alpha$ has recall at least $(1-2\alpha + \alpha \|q\|_1)/(1-\alpha)$. Thus, since $\alpha\in (\epsilon, 1-\epsilon)$, $\t \ge (1-2\alpha + \alpha \|q\|_1)/(1-\alpha) \ge \epsilon / (1-\epsilon)$ which implies that in our DP with accuracy parameter $\bar\epsilon$ it suffices to set $L_{\p} = (\frac{\epsilon}{1-\epsilon}) \cdot (1-\bar\epsilon)^{k-1} \ge (\frac{\epsilon}{1-\epsilon}) \cdot \exp(-\epsilon)$.
%= O((1-\frac{\epsilon}{2k})^k)$. 

\paragraph{Bounding $L_{\n}$.} 
For each $X\in \mathcal X$, let $\f_X$ denote the false positive rate of the optimal Equal Opportunity policy for group $X$. Similarly, let $\t_X$ denote the positive rate of (i.e., recall) the optimal policy $\pi^*$ for group $X\in \mathcal X$. By Equality of Opportunity property of $\pi^*$, $\t_{X} = \t$ for each $X\in {\mathcal X}$.
Next, we consider the following cases.

For any sufficiently small $\epsilon >0$, we need to set $L_{\n}$ so that by running the DP with accuracy parameter $\bar\epsilon$, we can approximate both true positive rate and false positive rate of the optimal Equal Opportunity policy within $(1-\epsilon)$-factor of their values. More precisely, we set $L_{\n}$ so that if for each group $X\in {\mathcal X}$ and any pair $(\t_X, \f_X)$ with $t_X \ge {L_{\p}}/(1-\epsilon/2)$, there exists a pair $(\bar{\t}_X, \bar{f}_X)$ such that $\bar{\t}_X \ge (1-\epsilon/2) \t_X$, $\bar{\f}_X \le \min(1, \max(L_{\n}, \f_X)/(1-\epsilon/2))$.
Finally, once the above property holds for all groups $X\in {\mathcal X}$, then for the corresponding policy $\pi$,  $\mathrm{precision}(\pi) > (1-\epsilon) \cdot \mathrm{precision}(\pi^*)$.

Let ${\mathcal X}_1 := \{X\in {\mathcal X} | \f_{X}/(1-\frac{\epsilon}{2}) \ge L_{\n}\}$ and ${\mathcal X}_2 := \{X\in {\mathcal X} | \f_{X}/(1-\frac{\epsilon}{2}) < L_{\n}\}$. Then,
\begin{align*}
    \frac{\mathrm{precision}(\pi)}{\mathrm{precision}(\pi^*)}
    &=\big(\frac{\|q\|_1 \cdot \bar{\t}}{\|q\|_1 \cdot \bar{\t} + \sum_{X\in \mathcal X} u_X \cdot \bar{\f}_X}\big) / \big(\frac{\|q\|_1 \cdot \t}{\|q\|_1 \cdot \t + \sum_{X\in {\mathcal X}}u_X \cdot \f_X}\big) \\
    &\ge \big(\frac{\|q\|_1 \cdot \bar{\t}}{\|q\|_1 \cdot \bar{\t} + \sum_{X\in {\mathcal X}_1} u_X \cdot \bar{\f}_X + \sum_{X\in {\mathcal X}_2} u_X \cdot \bar{\f}_X}\big) / \big(\frac{\|q\|_1 \cdot \t}{\|q\|_1 \cdot \t + \sum_{X\in {\mathcal X}_1}u_X \cdot \f_X}\big) \\
    &\ge \big(\frac{\|q\|_1 \cdot (1-\epsilon/2) \t}{\|q\|_1 \cdot \t + \sum_{X\in {\mathcal X}_1} \frac{u_X \cdot \f_X}{1-\epsilon/2} + \sum_{X\in {\mathcal X}_2} u_X \cdot L_{\n}}\big) / \big(\frac{\|q\|_1 \cdot \t}{\|q\|_1 \cdot \t + \sum_{X\in {\mathcal X}_1}u_X \cdot \f_X}\big)
\end{align*}
Next, we set $L_{\n}$ so that $\|q\|_1\cdot \t + \sum_{X\in {\mathcal X}_2} u_X L_{\n} \le \frac{\|q\|_1\cdot \t}{1-\frac{\epsilon}{2}}$. Since $\t \ge \epsilon / (1-\epsilon)$, it suffices to set $L_\n = \frac{\epsilon^2 \|q\|_1}{(2-\epsilon) (1-\epsilon) (1 - \|q\|_1)} = \Omega(\epsilon^2)$. Hence,
    \begin{align*}
        \big(\frac{\|q\|_1 \cdot (1-\frac{\epsilon}{2}) \t}{\|q\|_1 \cdot \t + \sum_{X\in {\mathcal X}_1} \frac{u_X \cdot \f_X}{1-\frac{\epsilon}{2}} + \sum_{X\in {\mathcal X}_2} u_X \cdot L_{\n}}\big) / \big(\frac{\|q\|_1 \cdot \t}{\|q\|_1 \cdot \t + \sum_{X\in {\mathcal X}_1}u_X \cdot \f_X}\big) \ge (1-\frac{\epsilon}{2})^2 > (1-\epsilon).
    \end{align*}

Finally, for each $X\in \mathcal X$, we run the DP algorithm for each group with accuracy parameter $\bar\epsilon$. 
By Lemma~\ref{lem:DP}, the DP algorithm finds a set $\{\t_X = (1-\bar\epsilon)^{i_X}, \f_x = (1-\bar\epsilon)^{j_X}\}_{X\in \mathcal X}$ (and a policy $\pi$ achieving these rates) where for each $X\in \mathcal X$, $\t_X \in [L_{\p}, 1], \f_X \in [L_{\n}, 1] $ such that
\begin{align*}
    \t_X = \t \ge (1-\epsilon/2)\cdot \t(\pi^*), \qquad\qquad
    \f_X \le \min\{1, \frac{\max\{L_{\n}, \f_X(\pi^*)\}}{1-\epsilon/2}\} &&\forall X\in \mathcal X,
\end{align*}
and for each $X\in \mathcal X$, $M_X[k, \t_X, \f_X] = \mathrm{true}$. Thus, by the bounds we just showed for the precision of such a policy, $\mathrm{precision}(\pi) \ge (1-\epsilon) \cdot \mathrm{precision}(\pi^*)$. Thus, $f_\alpha(\pi) \ge (1-\epsilon) \cdot f_\alpha(\pi^*)$.

As we need to run the DP algorithm for any of the $|\mathcal X|$ groups separately with the specified parameters $L_{\p}, L_{\n}$ and $\bar{\epsilon} = O(\epsilon/k)$, by Lemma~\ref{lem:DP-table-runtime}, the total time of the DP approach is $O(\frac{|{\mathcal X}| k \log^2(1/L_{\p}) \log^2(1/ L_{\n})}{\bar\epsilon^4}) 
= O(\frac{|{\mathcal X}| k^5 (\epsilon^2 + \log^2(1/\epsilon)) \log^2(1/\epsilon)}{\epsilon^4})
= O(\frac{|{\mathcal X}| k^5 \log^4(1/\epsilon)}{\epsilon^4})$.
\end{proof}

\begin{proof}[Proof of Theorem~\ref{thm:EQ-DP-min}]
First we show that in our setting, in order to guarantee $(1+\epsilon)$-approximations of recall and precision, it suffices to run the described DP and consider estimates of $\t$ (true positive rate) and $\f$ (false positive rate) of form $(1-\bar\epsilon)^i$ for $i\in \mathbb{N}$ in intervals $[L_{\p}, 1]$ and $[L_{\n}, 1]$ respectively, where $\bar\epsilon \le \epsilon/(2k)$. We provide tight bounds for $L_{\p}$ and $L_{\n}$. 

Note that since for any policy $\pi$, $\p_{i, \pi}, \n_{i, \pi}$ are non-decreasing in $i$, it suffices to provide ``large enough'' lowerbounds $L_{\p}$ and $L_{\n}$ for true positive rate and false positive rate in the final stage respectively (i.e., for $\t$ and $\f$). 

\paragraph{Bounding $L_{\p}$.} Consider the policy $\pi_{\bypass}$, which bypasses all the tests in both groups, i.e., $\pi^i_{X0} = \pi^i_{X1} = 1$ for all $i\in [k], X\in {\mathcal X}$. 
%Since $\pi_\bypass$ is an Equal Opportunity policy for the pipeline with $g_\alpha(\pi_\bypass) = 1-\alpha + \alpha/\|q\|_1$ and for any policy $\pi$ has $g_{\alpha}(\pi) \ge 1$, $\pi_\bypass$ is a $(1+\epsilon)$-approximation when $\alpha \le \frac{\epsilon \|q\|_1}{1- \|q\|_1}$.
%any optimal Equal Opportunity policy $\pi^*$ for $g_\alpha$ has recall at least $(1-\alpha)/(1-2\alpha + \alpha/\|q\|_1)$. 
Let $\tau_{\min} = \min_{X\in {\cal X}, j\in [k]} \tau^j_{X1}$. Then, by Theorem~\ref{thm:multi-eq-opp-prec}, Opportunity Ratio maximizes the precision and has recall at least $(\tau_{\min})^k$, in the optimal policy $\t \ge (\tau_{\min})^k$ which implies that in our DP with accuracy parameter $\bar\epsilon$ it suffices to set $L_{\p} = (\tau_{\min})^k \cdot (1-\bar\epsilon)^{k-1} \ge \exp(-\epsilon - k\ln (1/\tau_{\min}))$. 

\paragraph{Bounding $L_{\n}$.} 
For each $X\in \mathcal X$, let $\f_X$ denote the false positive rate of the optimal Equal Opportunity policy for group $X$. Similarly, let $\t_X$ denote the positive rate of (i.e., recall) the optimal policy $\pi^*$ for group $X\in \mathcal X$. By Equality of Opportunity property of $\pi^*$, $\t_{X} = \t$ for each $X\in {\mathcal X}$.
Next, we consider the following cases. 

For any sufficiently small $\epsilon >0$, we need to set $L_{\n}$ so that by running the DP with accuracy parameter $\bar\epsilon$, we can approximate both true positive rate and false positive rate of the optimal Equal Opportunity policy within $(1-\epsilon)$-factor of their values. More precisely, we set $L_{\n}$ so that if for each group $X\in {\mathcal X}$ and any pair $(\t_X, \f_X)$ with $t_X \ge {L_{\p}}$, there exists a pair $(\bar{\t}_X, \bar{f}_X)$ such that $\bar{\t}_X \ge (1-\epsilon/2) \t_X$, $\bar{\f}_X \le \min(1, \max(L_{\n}, \f_X)/(1-\epsilon/2))$.
Finally, once the above property holds for all groups $X\in {\mathcal X}$, then for the corresponding policy $\pi$,  $\mathrm{precision}(\pi) > (1-\epsilon) \cdot \mathrm{precision}(\pi^*)$.

Let ${\mathcal X}_1 := \{X\in {\mathcal X} | \f_{X}/(1-\epsilon/2) \ge L_{\n}\}$ and let ${\mathcal X}_2 := \{X\in {\mathcal X} | \f_{X}/(1-\epsilon/2) < L_{\n}\}$. Note that ${\mathcal X} = {\mathcal X}_1 \dot\cup {\mathcal X}_2$. Then,
\begin{align*}
    \frac{\mathrm{precision}(\pi)}{\mathrm{precision}(\pi^*)}
    &=\big(\frac{\|q\|_1 \cdot \bar{\t}}{\|q\|_1 \cdot \bar{\t} + \sum_{X\in \mathcal X} u_X \cdot \bar{\f}_X}\big) / \big(\frac{\|q\|_1 \cdot \t}{\|q\|_1 \cdot \t + \sum_{X\in {\mathcal X}}u_X \cdot \f_X}\big) \\
    &\ge \big(\frac{\|q\|_1 \cdot \bar{\t}}{\|q\|_1 \cdot \bar{\t} + \sum_{X\in {\mathcal X}_1} u_X \cdot \bar{\f}_X + \sum_{X\in {\mathcal X}_2} u_X \cdot \bar{\f}_X}\big) / \big(\frac{\|q\|_1 \cdot \t}{\|q\|_1 \cdot \t + \sum_{X\in {\mathcal X}_1}u_X \cdot \f_X}\big) \\
    &\ge \big(\frac{\|q\|_1 \cdot (1-\epsilon/2) \t}{\|q\|_1 \cdot \t + \sum_{X\in {\mathcal X}_1} \frac{u_X \cdot \f_X}{1-\epsilon/2} + \sum_{X\in {\mathcal X}_2} u_X \cdot L_{\n}}\big) / \big(\frac{\|q\|_1 \cdot \t}{\|q\|_1 \cdot \t + \sum_{X\in {\mathcal X}_1}u_X \cdot \f_X}\big)
\end{align*}
Next, we set $L_{\n}$ so that $\|q\|_1\cdot \t + \sum_{X\in {\mathcal X}_2} u_X L_{\n} \le \frac{\|q\|_1\cdot \t}{1-\frac{\epsilon}{2}}$. Since $\t \ge (\tau_{\min})^k$, it suffices to set $L_\n = \frac{\epsilon \|q\|_1\cdot (\tau_{\min})^k}{(2-\epsilon) (1-\|q\|_1)} = \Omega(\epsilon \cdot (\tau_{\min})^k)$. Hence,
\begin{align*}
    \big(\frac{\|q\|_1 \cdot (1-\epsilon/2) \t}{\|q\|_1 \cdot \t + \sum_{X\in {\mathcal X}_1} \frac{u_X \cdot \f_X}{1-\epsilon/2} + \sum_{X\in {\mathcal X}_2} u_X \cdot L_{\n}}\big) / \big(\frac{\|q\|_1 \cdot \t}{\|q\|_1 \cdot \t + \sum_{X\in {\mathcal X}_1}u_X \cdot \f_X}\big) \ge (1-\frac{\epsilon}{2})^2 > (1-\epsilon).
\end{align*}

Finally, for each $X\in \mathcal X$, we run the DP algorithm for each group with accuracy parameter $\bar\epsilon$. By Lemma~\ref{lem:DP}, the DP algorithm finds a set $\{\t_X = (1-\bar\epsilon)^{i_X}, \f_X = (1-\bar\epsilon)^{j_X}\}_{X\in \mathcal X}$ (and a policy $\pi$ corresponding to these values) where for each $X\in \mathcal X$, $\t_X \in [L_{\p}, 1], \f_X \in [L_{\n}, 1] $ such that
\begin{align}\label{eq:tpr-fpr-bounds}
    \t_X = \t \ge (1-\epsilon/2)\cdot \t(\pi^*), \quad
    \f_X \le \min(1, \frac{\max(L_{\n}, \f_X(\pi^*))}{1-\epsilon/2}) &&\forall X\in \mathcal X,
\end{align}
and for each $X\in \mathcal X$, $M_X[k, \t_X, \f_X] = \mathrm{true}$. Thus, by the bounds we just showed for the precision of such a policy, $1/\mathrm{precision}(\pi) \le (1+\epsilon) / \mathrm{precision}(\pi^*)$. Thus, $g_\alpha(\pi) \le (1+\epsilon) \cdot g_\alpha(\pi^*)$.

As we need to run the DP algorithm for any of the $|\mathcal X|$ groups separately with the specified parameters $L_{\p}, L_{\n}$ and $\bar{\epsilon} = O(\epsilon/k)$, by Lemma~\ref{lem:DP-table-runtime}, the total runtime is $O(\frac{|{\mathcal X}| k \log^2 (1/L_{\p}) \log^2 (1/L_{\n})}{\bar\epsilon^4}) = O(\frac{|{\mathcal X}| k^7 (\log^2(1/\epsilon) + k^2)}{\epsilon^4})$.
\end{proof}

%\section{Missing Proofs of Section \ref{sec-altmodels} \label{altsection-proofs}
%}

\section{Missing Proofs of Section \ref{eodds}}\label{altsection-proofs}
%First recall that the requirement of Equalized Odds mandates equal true positive and false positive rates for all groups, a stronger requirement than simply requiring Equal Opportunity. 
Similarly to Observation~\ref{obs:EOpp-cond}, we can show the following observation for the policies that satisfies the Equalized Odds requirement.
\begin{obsr}\label{obs:EOdd-cond}
For any policy $\pi$ that satisfies the Equalized Odds for a $k$-stage screening process with parameters $(\{\tau^i\}_{i\in [k]}, \{q_X, u_X\}_{X\in \mathcal X})$, there exists $M$ and $N$ such that for each $X\in \mathcal{X}$,
\begin{align*}
    M &:= \Pi_{i=1}^k \tau^i_{X1} \pi^i_{X1} + (1 - \tau^i_{X1}) \pi^i_{X0} , &&N := \Pi_{i=1}^k \tau^i_{X0} \pi^i_{X1} + (1 - \tau^i_{X0}) \pi^i_{X0}
\end{align*}
\end{obsr}
Note that as computed in Observation~\ref{obsr:interview-eff-formula}, for policy any satisfying the Equalized Odds, the interview efficiency of a policy $\pi$ for a $k$-stage process with parameters $(q, u, \{\tau^i\}_{i\in [k]})$ is $\frac{\|q\|_1 M}{\|q\|_1 M + \|u\|_1 N}$.

In the rest of the section and for the simplicity of the exposition, we assume there are exactly two groups in the population; $\mathcal X = \{A, B\}$. The result for the general setting can be derived similarly.

\begin{thm}\label{thm:ie-upbound}
The interview efficiency of any policy satisfying Equalized Odds for a single-stage screening process with parameters $(q, u, \tau)$ is at most $    \frac{q_A + q_B}{(q_A + q_B) + (u_A + u_B) \cdot \max(\frac{\tau_{A0}}{\tau_{A1}}, \frac{\tau_{B0}}{\tau_{B1}})}$.
% \begin{align*}
%     \frac{q_A + q_B}{(q_A + q_B) + (u_A + u_B) \cdot \max(\frac{\tau_{A0}}{\tau_{A1}}, \frac{\tau_{B0}}{\tau_{B1}})}
% \end{align*}
\end{thm}
\begin{proof}
Maximizing the interview efficiency, is equivalent to minimizing $N_{\tau, \pi} / M_{\tau, \pi}$; a minimizer of the inverse ratio is a maximizer of the interview efficiency and vice versa. Moreover, note that by the ``minimally effective'' property of the given test (i.e., Eq.~\eqref{eq:minimally-effective}), $N_{\tau, \pi} < M_{\tau, \pi}$. 
\begin{align*}
    &\frac{N_{\tau, \pi}}{M_{\tau, \pi}} = \frac{\tau_{A0} (\pi_{A1} - \pi_{A0}) + \pi_{A0}}{\tau_{A1} (\pi_{A1} - \pi_{A0}) + \pi_{A0}} \geq \frac{\tau_{A0} (\pi_{A1} - \pi_{A0})}{\tau_{A1} (\pi_{A1} - \pi_{A0})} = \frac{\tau_{A0}}{\tau_{A1}} 
    \text{ and } \\
    &\frac{N_{\tau, \pi}}{M_{\tau, \pi}} = \frac{\tau_{B0} (\pi_{B1} - \pi_{B0}) + \pi_{B0}}{\tau_{B1} (\pi_{B1} - \pi_{B0}) + \pi_{B0}} \geq \frac{\tau_{B0} (\pi_{B1} - \pi_{B0})}{\tau_{B1} (\pi_{B1} - \pi_{B0})} = \frac{\tau_{B0}}{\tau_{B1}}.
\end{align*}
In other words, $N_{\tau, \pi} \geq \max(\frac{\tau_{A0}}{\tau_{A1}}, \frac{\tau_{B0}}{\tau_{B1}}) \cdot M_{\tau, \pi}$. Hence, $\frac{(q_A + q_B) M_{\tau, \pi}}{(q_A + q_B) M_{\tau, \pi} + (u_A + u_B) N_{\tau, \pi}} \leq \frac{q_A + q_B}{q_A + q_B + (u_A + u_B) \cdot \max(\frac{\tau_{A0}}{\tau_{A1}}, \frac{\tau_{B0}}{\tau_{B1}})}$.
% \begin{align*}
%     \frac{(q_A + q_B) M_{\tau, \pi}}{(q_A + q_B) M_{\tau, \pi} + (u_A + u_B) N_{\tau, \pi}} 
%     &\leq \frac{q_A + q_B}{q_A + q_B + (u_A + u_B) \cdot \max(\frac{\tau_{A0}}{\tau_{A1}}, \frac{\tau_{B0}}{\tau_{B1}})}
% \end{align*}
\end{proof}

\begin{rem}
Note that we can generalize the result of Lemma~\ref{thm:ie-upbound} to a $k$-stage screening process with multiple groups $\mathcal{X}$. For any $j\in [k]$, let $\rho := \max_{X\in \mathcal{X}} \Pi_{j\in [k]}\frac{\tau_{X0}^j}{\tau_{X1}^j}$.
Any policy that satisfies Equalized Odds requirement at the end of the process (i.e., before the interview stage) has interview efficiency at most $\frac{\|q\|_1}{\|q\|_1 + \sum_{X\in \mathcal{X}}\rho u_X}$. 
To see this, note that similarly to the proof of Theorem~\ref{thm:ie-upbound} we can show that for every group $X\in \mathcal{X}$, $N_X \geq \rho \cdot M_X$.
\end{rem}

\begin{thm}\label{thm:Eodd-vs-Eopp}
Consider a $k$-stage screening process $(q, u, \tau)$ with multiple groups $\mathcal{X}$ whose tests are minimally effective. Let $\pi_{\mathrm{EOdd}}, \pi_{\mathrm{EOpp}}$ denote the interview efficiency maximizing policy that satisfies Equalized Odds and Equal Opportunity at the end of the process respectively. If $\max_{X\in \mathcal{X}} \Pi_{i\in [k]} \frac{\tau^i_{X0}}{\tau^i_{X1}} > \min_{X\in \mathcal{X}} \Pi_{i\in [k]} \frac{\tau^i_{X0}}{\tau^i_{X1}}$, then $\ie(q,u, \tau, \pi_{\mathrm{EOdd}}) < \ie(q,u, \tau, \pi_{\mathrm{EOpp}})$.

In particular, the gap between the interview efficiency of $\pi_{\mathrm{EOdd}}$ and $\pi_{\mathrm{EOpp}}$ can be as large as $\frac{1}{\|q\|_1}-\epsilon$ for any arbitrary $\epsilon >0$.\footnote{Note that the interview efficiency is always at most $1$ and the trivial Equalized Odds policy that bypasses all tests has interview efficiency $q$.}
\end{thm}
\begin{proof}
The proof of the first part directly follows from the interview efficiency of opportunity ratio policy (Theorem~\ref{thm:multi-eq-opp-prec}) and the upper bound for the interview efficiency of Equalized Odds policies (Theorem~\ref{thm:ie-upbound})

For the second part, consider a pipeline in which there exists a $X^*\in \mathcal{X}$ such that for every $X\in \mathcal{X}\setminus X^*$, $\Pi_{i\in [k]} \frac{\tau^i_{X0}}{\tau^i_{X1}} =0$ and $\Pi_{i\in [k]} \frac{\tau^i_{X^*0}}{\tau^i_{X^*1}} =(1-\delta)^k$. Further, for every $X\in \mathcal{X}\setminus X^*$, let $q_{X} = \frac{\gamma}{k}, u_{X} = \frac{1-\gamma-\mu}{k-1}$ and $q_{X^*} = \frac{\gamma}{k}, u_{X^*} = \mu$. Then, it is straightforward to check that $\ie(\pi_{\mathrm{EOpp}}) = \frac{\gamma}{\gamma + \mu\cdot(1-\delta)^k}$ and $\ie(\pi_{\mathrm{EOdd}}) = \frac{\gamma}{\gamma + (1-\gamma)\cdot (1-\delta)^k}$. As we set $\delta, \mu$ to sufficiently small values, $\ie(\pi_{\mathrm{EOpp}})/\ie(\pi_{\mathrm{EOpp}}) = 1/\gamma -\epsilon = 1/\|q\|_1 -\epsilon$.
\end{proof}

%\avnote{We can keep the following is the section is moved to the appendix. Otherwise, it is not interesting enough to be in the main body.}

%In particular, the above remark shows that if we want to satisfy the Equalized Odds, the optimal policy cannot achieve a guarantee better than a policy that in each stage matches the accuracy of the worst performance of the given test over groups $A$ and $B$ (i.e., $\max(\tau_{A0}/\tau_{A1}, \tau_{B0}/ \tau_{B1})$.  

Next, we show the following structure on a non-trivial optimal solution (i.e., one maximizing the interview efficiency). Note that $\pi= \boldsymbol{1}$ or $\pi=\boldsymbol{0}$ are the two trivial solutions satisfying the Equalized Odds for any given test.
\begin{obsr}\label{obr:Eodd-opt-structure}
For any pipeline $(\tau, q,u)$, in any {\em non-trivial} optimal policy $\pi$, $\min(\pi_{A1}, \pi_{A0}, \pi_{B1}, \pi_{B0}) =0$. Moreover, there exists an optimal policy such that $\max(\pi_{A1}, \pi_{A0}, \pi_{B1}, \pi_{B0}) =1$.
\end{obsr}
\begin{proof}
First, note that by the ``minimally effective'' property of the given test (i.e., Eq.~\eqref{eq:minimally-effective}), $N_{\tau, \pi} < M_{\tau, \pi}$.
Suppose that $\min(\pi_{A1}, \pi_{A0}, \pi_{B1}, \pi_{B0}) =\epsilon$. This implies that $M_{\tau,\pi} > N_{\tau, \pi} \ge \epsilon$ 
Then, by subtracting $\epsilon$ from all $\pi$ values, the new policy still satisfies the Equalized Odds and it only increases the interview efficiency. Formally, for $\epsilon>0$
\begin{align*}
    \frac{\|q\|_1 \cdot M_{\tau, \pi}}{\|q\|_1 \cdot M_{\tau, \pi} + \|u\|_1\cdot N_{\tau, \pi}} < \frac{\|q\|_1 \cdot (M_{\tau, \pi} - \epsilon)}{\|q\|_1 \cdot (M_{\tau, \pi} - \epsilon) + \|u\|_1\cdot (N_{\tau, \pi} - \epsilon)}
\end{align*}
The above inequality holds since
\begin{align*}
    &N_{\tau, \pi} < M_{\tau, \pi} \\
    \Rightarrow &-\|u\|_1\epsilon N_{\tau, \pi} > - \|u\|_1\epsilon M_{\tau, \pi} 
     \\
    \Rightarrow &(\|q\|_1 M_{\tau, \pi}^2 - \|q\|_1 \epsilon M_{\tau, \pi} + \|u\|_1 M_{\tau, \pi} N_{\tau, \pi}) - \|u\|_1 \epsilon N_{\tau, \pi} 
    > (\|q\|_1 M_{\tau, \pi}^2 - \|q\|_1\epsilon M_{\tau, \pi} + \|u\|_1 M_{\tau, \pi} N_{\tau, \pi}) - \|u\|_1 \epsilon M_{\tau, \pi} \\
    \Rightarrow & M_{\tau, \pi} (\|q\|_1 M_{\tau, \pi} + \|u\|_1 N_{\tau, \pi})
     - \epsilon (\|q\|_1 M_{\tau, \pi} + \|u\|_1 N_{\tau, \pi}) 
    > M_{\tau, \pi}(\|q\|_1 (M_{\tau, \pi} - \epsilon) + \|u\|_1 (N_{\tau, \pi} - \epsilon)) \\
    \Rightarrow & \frac{M_{\tau, \pi} - \epsilon}{\|q\|_1(M_{\tau, \pi} - \epsilon) + \|u\|_1 (N_{\tau, \pi} - \epsilon)}  
    > \frac{M_{\tau, \pi}}{\|q\|_1 M_{\tau, \pi} + \|u\|_1 N_{\tau, \pi}} \quad\rhd\text{since $\|q\|_1 (M_{\tau, \pi} - \epsilon) + \|u\|_1  (N_{\tau, \pi} - \epsilon) > 0$}
\end{align*}
In particular, this implies that in any optimal policy, $\min(\pi_{A1}, \pi_{A0}, \pi_{B1}, \pi_{B0}) = 0$.

The second part of the statement follows simply from the fact that if we multiply all $\pi$ values by a constant $c>1$ so that they remain feasible (i.e., none of $\pi$ values goes above one), the interview efficiency of the policy $c\pi$ and the policy $\pi$ are the same. 
\end{proof}

Note that though it seems counter-intuitive, it might be the case $\pi_{A0} = \argmax(\pi_{A1}, \pi_{A0}, \pi_{B1}, \pi_{B0})$ and/or $\pi_{A1} = \argmin(\pi_{A1}, \pi_{A0}, \pi_{B1}, \pi_{B0})$.

%% file: appendix.tex
\section{An FPTAS Algorithm for Screening Processes with Same Policy for All Groups} \label{sec:single-policy-app}
% The first, and, of practical relevance is ensuring fair behavior when there are \textit{not} multiple classifiers fine-tuned for each group. 
% While directly incorporating the the demographic feature into the data representation has been proposed in the fairness in machine learning literature \cite{dwork2012fairness}, in alternative words, this could mean giving two individuals of different demographic groups  different predictions or decisions even if 
% every other feature in their corresponding data point is identical. In some regulatory regimes, this fairness-through-awareness may be illegal or problematic, even when intended to ensure equitable treatment. 

% In our setting,  if we are constrained to follow \textit{group-blindness}, there be would only one set of tests and one ordering of the tests that all applicants are tested on. Analogously to the previous setting, the action space of algorithm remains modifying the promotion probabilities, but we now only have one set of policies to modify. 

Here, we devise a slightly different DP algorithm. Instead of running the DP algorithm for each group separately (as in Section~\ref{sec:fptas}), we run a single DP algorithm for all groups simultaneously. Hence, all policies $\{\pi_X\}_{X \in {\mathcal X}}$ are the same. In our DP approach, we use the same discretization technique and only consider powers of $(1-\epsilon)$.

\paragraph{Solving the DP} 
Consider the first level, $i=1$.
For any given parameters $\{j_{X,0}, j_{X, 1}\}_{X\in \mathcal X}$, where for each group $X\in \mathcal X$, $j_{X,0}\in [0, L_{\n}]$ and $j_{X,1} \in [0, L_{\p}]$, $M[1,\{j_{X0}, j_{X1}\}_{X\in {\mathcal X}}]=\mathrm{true}$ iff the following has a feasible solution.
\begin{align}\label{eq:base-rule-single} 
    \tau^1_{X0} x + (1- \tau^1_{X0}) y &\leq (1-\epsilon)^{j_{X0}}, 
    &\tau^1_{X1} x + (1- \tau^1_{X1}) y \geq (1-\epsilon)^{j_{X1}} &&\forall X\in {\mathcal X}
\end{align}

Next, we describe the update rule for $i>1$. For any $X\in \mathcal X$, $\n_X \in [0, \ell_\n]$ and $\p_X\in [0, \ell_\p]$,
\begin{align*}
    M[i+1, \{\p_X, \n_X\}_{X \in {\mathcal X}}] &=  \bigvee_{\{j_{X1}, j_{X0}\}_{X\in {\mathcal X}} \in {\mathcal F}_{i+1}} M[i, \{{\p_X}-{j_{X1}},  {\n_X}-{j_{X0}}\}_{X\in {\mathcal X}}]
\end{align*}
where ${\mathcal F}_{i+1}$ is the set of $\{j_{X1} \le \p_X, j_{X0}\le \n_X\}_{X\in {\mathcal X}}$ for which the following system of linear inequalities has a feasible solution,
\begin{align}\label{eq:update-rule-single}
    \tau^{i+1}_{X1} x+ (1-\tau^{i+1}_{X1}) y &\ge (1-\epsilon)^{j_{X1}},
    &\tau^{i+1}_{X0} x+ (1-\tau^{i+1}_{X0}) y \le (1-\epsilon)^{j_{X0}} &&\forall X\in {\mathcal X}
\end{align}
\begin{lem}\label{lem:single-policy-DP}
For any $i\in [k]$, if there exists an Equal Opportunity policy $\pi$ treating all groups similarly, with true positive rate $\t_{X,i} \ge L_{\p}/(1-\epsilon)^{i-1}$, false positive rate $\f_{X,i}$ for $X\in {\mathcal X}$, then there exist $\{j_{X1}, j_{X0}\}_{X\in {\mathcal X}}$ such that $M[i, \{j_{X1}, j_{X0}\}_{X\in {\mathcal X}}] = \mathrm{true}$, where for each $X\in {\mathcal X}$, $(1-\epsilon)^{j_{X1}} \ge \t_{X,i} \cdot (1-\epsilon)^{i-1}$ and $(1-\epsilon)^{j_{X0}} \le \min\{ 1, \max\{ L_{\n}, \f_{X,i}\}/ (1-\epsilon)^{i-1}\}$.

In other words, if the policy $\pi$ exists then the DP approach finds a policy with true positive rate at least $(1-\epsilon)^{j_{X1}}$ and false positive rate at most $(1-\epsilon)^{j_{X0}}$ for each $X\in \mathcal X$.
\end{lem}

\begin{proof}
The proof is by induction. For the base case ($i=1$), let $\t_{X,1}$ and $\f_{X,1}$ denote the true positive rate and the false positive rate of $\pi$ by the end of level $1$ for each group $X\in \mathcal X$. 
The existence of $\pi$ guarantees that the system of inequalities Eq.~\eqref{eq:base-rule-single} with $(j_{X0} = \lfloor \log_{1-\epsilon} \f_{X,1} \rfloor, j_{X1} = \lceil \log_{1-\epsilon}\t_{X,1} \rceil \leq \ell_{\p})$ has a feasible solution.  
More precisely, by setting $(x_X = \pi_{X1}, y_X = \pi_{X0}) \forall X\in \mathcal X$, 
\begin{align*} 
    &\tau^1_{X0} x_X + (1- \tau^1_{X0}) y_X = \f_{X1} \le (1-\epsilon)^{\lfloor \log_{1-\epsilon} \f_{X1} \rfloor} = (1-\epsilon)^{j_{X0}}, &&\forall X\in \mathcal X \\ 
    &\tau^1_{X1} x_X + (1- \tau^1_{X1}) y_X = \t_{X1} \geq (1-\epsilon)^{\lceil \log_{1-\epsilon} \t_{X1} \rceil} = (1-\epsilon)^{j_{X1}}&&\forall X\in \mathcal X
\end{align*}
Next, we consider $i>1$ and we assume that the claim holds for all values $i' < i$. For each $X\in \mathcal X$, let $M_{X,i} := \tau^i_{X1}\pi^i_{X1} + (1 - \tau^i_{X1}) \pi^i_{X0}$ and $N_{X,i} := \tau^i_{X0}\pi^i_{X1} + (1-\tau^i_{X0})\pi^i_{X0}$. Note that for each $X\in \mathcal X$, $\t_{X,i} = \t_{X,i-1} \cdot M_{X,i}$ and $\f_{X,i} = \f_{X,i-1} \cdot N_{X,i}$.

By the induction hypothesis and considering the first $i-1$ levels in the pipeline, since $\t_{X,i-1} \ge \t_{X, i} \ge L_{\p}/(1-\epsilon)^{i-1} > L_{\p}/(1-\epsilon)^{i-2}$ and $\f_{X,i-1} \ge \f_{X, i}$, there exist $j'_{X1}\in[0, L_{\p}]$ and $j'_{X0}\in [0, L_{\n}]$ such that $M[i-1, \{j'_{X1}, j'_{X0}\}_{X\in \mathcal X}] = \mathrm{true}$ and $(1-\epsilon)^{j'_{X1}} \ge \t_{X,i-1} \cdot (1-\epsilon)^{i-2}$ and $(1-\epsilon)^{j'_{X0}} \le \min\{1, \max\{L_{\n}, \f_{X,i-1}\}/(1-\epsilon)^{i-2}\}$. 
More precisely, the algorithm finds a policy $\bar{\pi}$ with true positive rate at least $(1-\epsilon)^{j'_{X1}}$ and false positive rate at most $(1-\epsilon)^{j'_{X0}}$ for each $X\in \mathcal X$.

Next, for each $X\in \mathcal X$, by setting $(\bar{\pi}^i_{X1} = \pi^i_{X1}, \bar{\pi}^i_{X0} = \pi^i_{X0})$ and $(j_{X1} :=\argmin_{j} \{(1-\epsilon)^j \le \t_{X,i}(\bar\pi)\}, j_{X0} := \argmax_{j} \{(1-\epsilon)^j \ge \f_{X,i}(\bar\pi)\})$,  
\begin{align*}
    (1-\epsilon)^{j_{X1}} > (1-\epsilon)\cdot \t_{X,i}(\bar \pi)
    &= (1-\epsilon)\cdot \t_{X,i-1}(\bar \pi) \cdot M_{X,i} &&\rhd\text{by definition of $j_{X1}$}\\
    &\ge (1-\epsilon)\cdot (1-\epsilon)^{j'_{X1}} \cdot M_{X,i} &&\rhd\text{by $\t_{X, i-1}(\bar\pi) \ge (1-\epsilon)^{j'_{X1}}$}\\ 
    &\ge \t_{X, i-1} \cdot (1-\epsilon)^{X, i-1} \cdot M_{X, i} &&\rhd\text{by induction hypothesis}\\
    &= \t_{X, i} \cdot (1-\epsilon)^{i-1}. 
\end{align*}
Similarly,
\begin{align*}
    (1-\epsilon)^{j_{X0}} < \min\{1, \frac{\f_{X,i}(\bar \pi)}{1-\epsilon}\} 
    &= \min\{1, N_{X,i} \cdot \frac{\f_{X, i-1}(\bar \pi)}{1-\epsilon}\} &&\rhd\text{by definition of $j_{X0}$}\\
    &\le \min\{1, (1-\epsilon)^{j'_{X0}} \cdot \frac{N_{X,i}}{1-\epsilon}\} &&\rhd\text{by $\f_{X, i-1}(\bar\pi) \le (1-\epsilon)^{j'_{X0}}$} \\
    &\le \min\{1, \frac{\max\{L_{\n}, \f_{X, i-1}\}}{(1-\epsilon)^{i-2}} \cdot \frac{N_{X,i}}{1-\epsilon}\} &&\rhd\text{by induction hypothesis}\\ 
    &\le \min\{1, \frac{\max\{L_{\n}, \f_{X,i}\}}{(1-\epsilon)^{i-1}}\}
\end{align*}
which completes the proof.
\end{proof}

\begin{lem}\label{lem:single-policy-DP-runtime}
For an accuracy parameter $\epsilon$ and lowerbounds on the false positive rate, $L_{\n}$, and the true positive rate, $L_{\p}$, the (single policy) DP algorithm runs in time $O(\frac{k \log^{2|{\mathcal X}|}(1/L_{\p}) \log^{2|{\mathcal X}|}(1/L_{\n})}{\epsilon^{4|{\mathcal X}|}})$ and finds a policy $\pi$ with true positive rate at least $(1-\epsilon)^{k-1} \cdot \t_X$ and false positive rate at most $\min\{ 1, \max\{ L_{\n}, \f_X\}/ (1-\epsilon)^{k-1}\}$ for each $X\in \mathcal X$.
\end{lem}
\begin{proof}
The size of table is $O(k \ell_{\p}^{|{\mathcal X}|} \ell_{\n}^{|{\mathcal X}|})$ and updating each entry in the table takes $O(\ell_\p^{|{\mathcal X}|} \ell_\n^{|{\mathcal X}|})$. Hence, the total runtime to compute all entries in the DP table is $O(k\ell^{2|{\mathcal X}|}_\p \ell^{2|{\mathcal X}|}_\n) = O(\frac{k \log^{2|{\mathcal X}|}(1/L_{\p}) \log^{2|{\mathcal X}|}(1/L_{\n})}{\epsilon^{4|{\mathcal X}|}})$.

Now we apply the DP approach and by Lemma~\ref{lem:single-policy-DP}, the solution returned by the algorithm has the true positive rate and the false positive rate satisfying the guarantee of the statement.
\end{proof}
\paragraph{Implications of DP} Here, similarly to Section~\ref{sec:fptas}, we present FPTAS algorithms for the single policy setting with various pipeline efficiency objective using the modified DP approach described above when the number of different protected groups in the population is a fixed constant; $|\mathcal{X}| = O(1)$. 
\begin{thm}\label{thm:EQ-DP-max-single}
Consider a $k$-stage screening process with parameters $(u, q, \tau, \mathcal X)$ where $|\mathcal{X}| = O(1)$. For any policy $\pi$, let $f_{\alpha}(\pi) = \mathrm{recall}(\pi) + \alpha \cdot \mathrm{precision}(\pi)$ where $\alpha > 0$. Given an accuracy parameter $\epsilon$, there exists an FPTAS that runs in time $O(\frac{k^{4|{\mathcal X}|} \log^{2|{\mathcal X}|}(1/\epsilon)}{\epsilon^{4|{\mathcal X}|}})$ and finds an Equal Opportunity policy $\pi$ treating all groups similarly such that $f_{\alpha}(\pi) \ge (1-\epsilon)f_{\alpha}(\pi^*)$ where $\pi^*$ maximizes $f_\alpha$ over Equal Opportunity policies treating all groups similarly. 
\end{thm}
\begin{thm}\label{thm:EQ-DP-min-single}
Consider a $k$-stage screening process with parameters $(u, q, \tau, \mathcal X)$ where $|\mathcal{X}| = O(1)$. For any policy $\pi$, let $g_{\alpha}(\pi) = 1/\mathrm{recall}(\pi) + \alpha / \mathrm{precision}(\pi)$ where $\alpha > 0$. Given an accuracy parameter $\epsilon$, there exists an FPTAS that runs in time $O(\frac{k^{4|{\mathcal X}|} \log^{2|{\mathcal X}|}(1/\epsilon)}{\epsilon^{4|{\mathcal X}|}})$ and finds an Equal Opportunity policy $\pi$ treating all groups similarly such that $g_{\alpha}(\pi) \le (1+\epsilon)g_{\alpha}(\pi^*)$ where $\pi^*$ minimizes $g_\alpha$ over Equal Opportunity policies treating all groups similarly.
\end{thm}
The proof of above theorems are identical to Theorem~\ref{thm:EQ-DP-max} and Theorem~\ref{thm:EQ-DP-min}.

\section{Additional details in Linear Combination Counter Examples \label{appendix-linear-combination-counter}}
%\avnote{Revise this section. My suggestion: (1) Before F.1., state the goal of this section. (2) Add a theorem statement and prove it formally (if the proof has multiple part, instead of having subsections, use lemmas.)}
In this section, we show that one cannot ``locally score'' tests when determining the optimum policy (the policy that maximizes a linear combination of precision and recall).  Specifically, we give a setting with three levels of tests $t_1, t_2, t_3$ such that if only the first two levels $t_1$ and $t_2$ are available, then the optimal solution is to use $t_1$ and bypass $t_2$, but if $t_3$ is also available then the optimal solution is to bypass $t_1$ and use $t_2$ and $t_3$.  Therefore, the question of how to best use two tests may depend on what tests are available at other levels.
Note that in this example there is only one group and we do not have fairness constraints. 

First, we show the following useful property of optimal policies for a pipeline where the first level has test statistics $(1/2, 0)$ and all other levels have test statistics $(1-\delta, 1/2)$.
\begin{lem}\label{lem:prop-single}
In any $k$-stage pipeline where the first stage has test statistics $(1/2, 0)$ and the rest of the stages have tests with statistics $(1- \delta, 1/2)$, the optimal policy is of the form $(1, \pi^1_0), \cdots, (1, \pi^k_0)$.
% \begin{itemize}
%     \item $(1, \pi^1_0), \cdots, (1, \pi^k_0)$ where $\pi^1_0 >0$, or
%     \item $(1, 0), (1, 1)\cdots, (1,1)$, i.e., fully use the first test and bypass the rest.
% \end{itemize}
\end{lem}
\begin{proof}
By Lemma~\ref{lem:pi-cond}, if the False Positive rate is non-zero, in the optimal policy, for every $i\in [k]$, $(1 - \pi^i_1) \pi^i_0 = 0$. Next, we show that in this setting with only one group, for every $i\in [k]$, $\pi^i_1 = 1$. Suppose that there exists a level $i\in [k]$ such that $\pi^i_0 = 0$. Then, if $\pi^i_1 <1$, by increasing $\pi^i_1$ to $1$, the True Positive rate and False Positive rate increase by the same factor. Therefore, the precision remains unchanged and the recall increases; hence, the pipeline efficiency strictly increases.

Next, we consider the case where the optimal policy has precision one (i.e., its False Positive is zero). In any such policy, $\pi^1_0 = 0$. Note that once the precision is $1$, the optimal policy maximizes recall. Hence, the optimal policy is to fully use $t_1$ ($\pi^1_1 =1, \pi^1_0 =0$) and bypass the rest of tests (for every $1<i\le k$, $\pi^i_1 = \pi^i_0 = 1$). 
\end{proof}

\begin{thm}
\label{thmlinearexample}
When the objective is to maximize a linear combination of precision and recall 
%with combinations weights $\alpha, \beta > 0 $ 
in a multi-stage screening process, there exist test parameters $T$ and base rate $p$ such that the maximal score policy 
%for some of the tests 
switches 
%(e.g. two tests $j$ and $k$ with $\pi_{j} = (1,0)$ and $\pi_{k} = (1,1)$) 
when more tests become available.  Specifically, when only tests $t_1$ and $t_2$ are available, the optimal policy is to use $t_1$ and bypass $t_2$ $((1,0), (1,1))$, but if test $t_3$ is also available, the optimal policy is to bypass $t_1$ and use $t_2$ and $t_3$ $((1,1), (1,0), (1,0))$.
\end{thm}
\begin{proof}
Consider base rate $p= P(x=1) = 1/2$ and test $t_1=(\tau_1, \tau_0) = (1/2,0)$ and test $t_2 =t_3= (1-\delta, 1/2)$. 
Let $\delta=\frac{1}{100}$. The linear objective function is $f(\pi) = \mathrm{recall}(\pi) + 2 \cdot \mathrm{precision}(\pi)$. 
Next, we consider two cases: (1) $k=2$ and (2) $k=3$.
\paragraph{Case 1: Two test ($k=2$).} By Lemma~\ref{lem:prop-single}, the optimal policy is of form $((1, \pi^1_0), (1, \pi^2_0))$. By numerical analysis\footnote{Using WolframAlpha.}, the local optimum policies (w.r.t.~$f$) are $((1,0), (1,1))$ and $((1,1), (1,0))$. Next, we compute the score of these two policies: $f((1,0), (1,1)) = 2.5$ and $f((1,1), (1,0)) < 2.32$. Hence, in this case, the optimal policy is to fully use $t_1$ and bypass $t_2$, i.e., $((1,0), (1,1))$.

% either to fully use $t_1$ and bypass $t_2$ (i.e., $((1, 0), (1, 1))$) or to pick $((1, \pi^1_0), (1, \pi^2_0))$ where $\pi^1_0 >0$. 
%\AB{[AB: Isn't the first option a special case of the second option?  Or should $\pi_0^1 =1$ here?]} \AV{Sure, I added the condition $\pi^1_0 > 0$ to make the cases disjoint.}
% \begin{itemize}
%     \item {\bf $((1, 0), (1, 1))$.} $f((1, 0), (1, 1)) = 1/2 + 2\cdot 1 = 5/2$.
%     \item {\bf $((1, \pi^1_0), (1, \pi^2_0))$ where $\pi^1_0 >0$.} By numerical analysis\footnote{Using WolframAlpha.}, the maximizer of $f$ of this form is obtained by the policy $((1,1), (1,0))$ and $f((1,1), (1,0))<2.32$. 
%     %\AB{[AB: According to my calculations, this policy has recall $=1-\delta$ and precision $=\frac{1-\delta}{(1-\delta)+1/2} < 2/3$, so its overall score is less than $2 + 1/3$, which is smaller than the above $2 + 1/2$.]} \AV{You are right. I fixed it.}
% \end{itemize}
\paragraph{Case 2: Three tests ($k=3$).} Similarly to the previous case, the optimal policy for the given pipeline efficiency objective is of form $((1, \pi^1_0), (1, \pi^2_0), (1, \pi^3_0))$.
By numerical analysis, the local optimum policies (w.r.t.~$f$) are $((1,0), (1,1), (1,1))$ and $((1,1), (1,0), (1,0))$. Next, we compute the score of these two policies: $f((1,0), (1,1), (1,1)) = 2.5$ and $((1,1), (1,0), (1,0)) > 2.57$. This time, the optimal policy is to bypass $t_1$ and fully use $t_2, t_3$, i.e. $((1,1), (1,0), (1,0))$.

% \begin{itemize}
%     \item {\bf $((1, 0), (1, 1), (1,1))$.} $f((1, 0), (1, 1), (1,1)) = 1/2 + 2\cdot 1 = 5/2$.
%     \item {\bf $((1, \pi^1_0), (1, \pi^2_0), (1, \pi^3_0))$ where $\pi^1_0 >0$.} By numerical analysis, the maximizer of $f$ of this form is obtained by the policy $((1,1), (1,0), (1,1))$ and $f((1,1), (1,0), (1,1))>2.57$. 
% \end{itemize}

Therefore, while in the first setting ({\em only $t_1$ and $t_2$ are available}) the optimal policy is to fully use $t_1$ and bypass $t_2$, once $t_3$ becomes available, the optimal policy changes to bypass $t_1$ and fully use $t_2$ and $t_3$.  
\end{proof}